\documentclass[10pt,twocolumn,letterpaper]{article}

\usepackage[pagenumbers]{cvpr} % To force page numbers, e.g. for an arXiv version

\usepackage[dvipsnames]{xcolor}

\usepackage{url}
\usepackage{epsfig}
\usepackage{graphicx}
\usepackage{amsmath}
\usepackage{amssymb}
\usepackage{xspace}
\usepackage{amstext}
\usepackage{utfsym}
\usepackage{algorithm}
\usepackage{algpseudocode}

\usepackage{booktabs}
\usepackage{multirow}

\usepackage{newfloat}
\usepackage{listings}
\usepackage{bibentry}
\usepackage{bm}
\definecolor{dark-red}{rgb}{0.4,0.15,0.15}
\definecolor{dark-blue}{rgb}{0.15,0.15,0.4}
\definecolor{medium-blue}{rgb}{0,0,0.5}
\definecolor{cvprblue}{rgb}{0.21,0.49,0.74}

\usepackage[pagebackref,breaklinks,colorlinks,citecolor=cvprblue]{hyperref}

\newcommand{\rmnum}[1]{\romannumeral #1}
\newcommand{\Rmnum}[1]{\expandafter\@slowromancap\romannumeral #1@}

\usepackage[accsupp]{axessibility}
\newcommand\blfootnote[1]{%
  \begingroup
  \renewcommand\thefootnote{}\footnote{#1}%
  \addtocounter{footnote}{-1}%
  \endgroup
}

\def\paperID{4662} % *** Enter the Paper ID here
\def\confName{CVPR}
\def\confYear{2024}

\title{De-confounded Data-free Knowledge Distillation for Handling Distribution Shifts}

\author{Yuzheng Wang$^{1  *}$ $\quad$
        Dingkang Yang$^{1  *}$ $\quad$
        Zhaoyu Chen$^{1}$ $\quad$
        Yang Liu$^{1}\quad$
        Siao Liu$^{1}\quad$ \\
        Wenqiang Zhang$^{2}\quad$ 
        Lihua Zhang$^{1 \dagger}\quad$
        Lizhe Qi$^{1,2,3 \dagger}$ \\ 
        \small$^1$Shanghai Engineering Research Center of AI \& Robotics, Academy for Engineering \& Technology, Fudan University \\
        \small$^2$Engineering Research Center of AI \& Robotics, Ministry of Education, Academy for Engineering \& Technology, Fudan University \\
        \small$^3$Green Ecological Smart Technology School-Enterprise Joint Research Center \\
        {\tt\small \{yzwang20, dkyang20\}@fudan.edu.cn}
}

\begin{document}

\maketitle

\blfootnote{$^{*}$Equal contribution.\ \ $^\dagger$Corresponding authors.}

\begin{abstract}

Data-Free Knowledge Distillation (DFKD) is a promising task to train high-performance small models to enhance actual deployment without relying on the original training data.
Existing methods commonly avoid relying on private data by utilizing synthetic or sampled data.
However, a long-overlooked issue is that the severe distribution shifts between their substitution and original data, which manifests as huge differences in the quality of images and class proportions.
The harmful shifts are essentially the confounder that significantly causes performance bottlenecks.
To tackle the issue, this paper proposes a novel perspective with causal inference to disentangle the student models from the impact of such shifts.
By designing a customized causal graph, we first reveal the causalities among the variables in the DFKD task.
Subsequently, we propose a Knowledge Distillation Causal Intervention (KDCI) framework based on the backdoor adjustment to de-confound the confounder.
KDCI can be flexibly combined with most existing state-of-the-art baselines. 
Experiments in combination with six representative DFKD methods demonstrate the effectiveness of our KDCI, which can obviously help existing methods under almost all settings, \textit{e.g.}, improving the baseline by up to 15.54\% accuracy on the CIFAR-100 dataset.

\end{abstract}

\section{Introduction}

Deep Neural Networks (DNNs), as a powerful and reliable tool, are increasingly expected to be applied to practical artificial intelligence scenes \cite{he2016deep,liu2021swin,yang2023target,liu2023stochastic,yang2024robust,yang2022disentangled,yang2022learning}.
Despite significant progress, good performance of deep learning models is often inseparable from large-scale models \cite{devlin2018bert,karras2019style,brown2020language,yang2024how2comm,chen2022towards,liu2023amp} and high-quality original training data \cite{radford2021learning,ramesh2022hierarchical,deng2009imagenet,yang2023aide,wang2023adversarial,ge2022zoom,liu2023improving}.
The dependencies hinder the deployment of this technology on mobile devices and data privacy scenes.
Therefore, model compression and data-free technology have become the key to breaking through the bottleneck.
To this end, Lopes \textit{et al.} \cite{lopes2017data} propose the Data-Free Knowledge Distillation (DFKD) task.
In this process, knowledge is transferred from the cumbersome
model to a small model that is more suitable for deployment \cite{hinton2015distilling,romero2014fitnets,wang2024out} without relying on the original training data.
As a result, DFKD has received more attention due to its convenience and wide application.

\begin{figure}[t]
	\centering
	\includegraphics[scale=0.32]{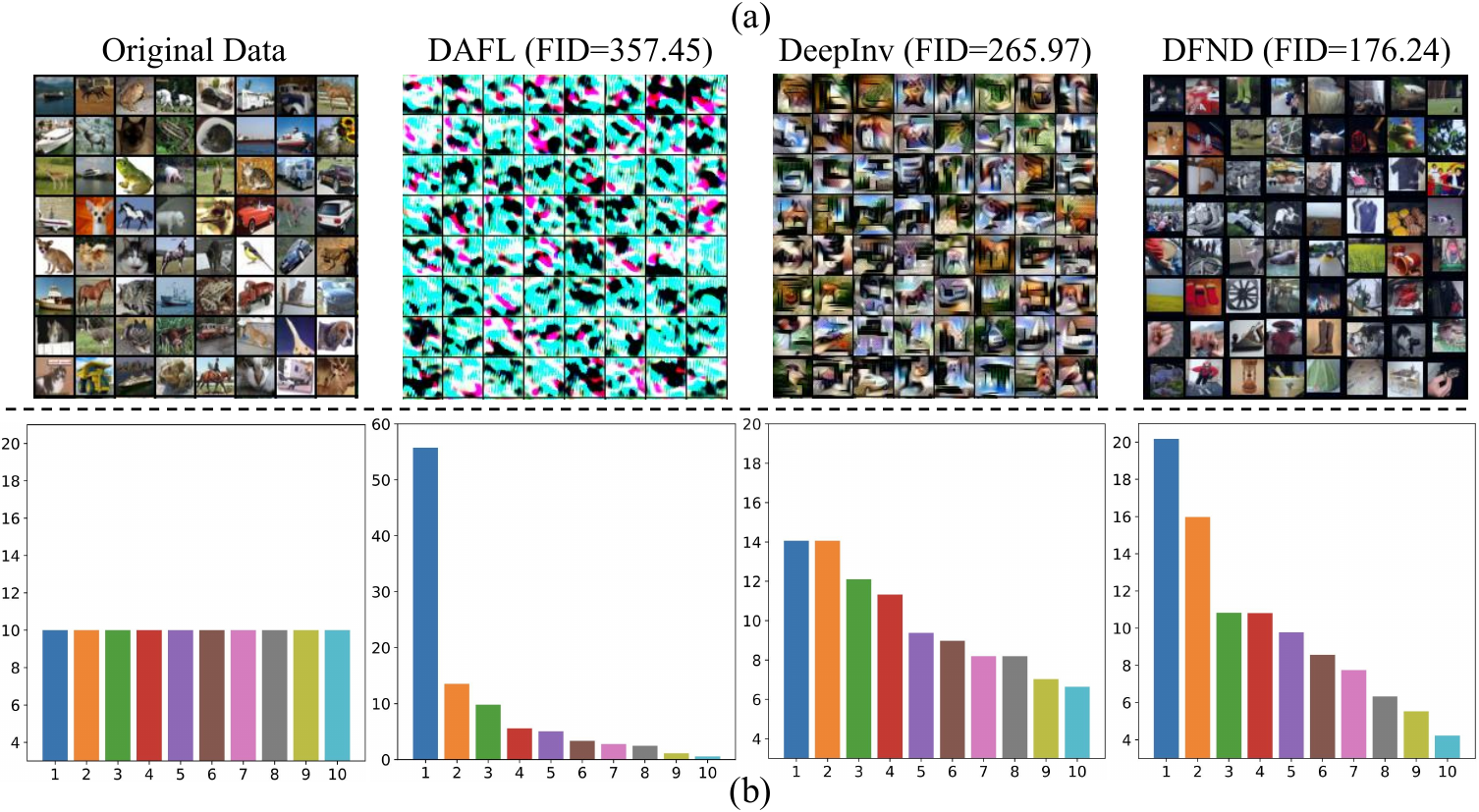}
	\vspace{-0.35cm}
	\caption{Diagrams of the distribution shifts between the original and substitute data for existing DFKD methods on CIFAR-10. (a) represents the random visualization and FID score of the synthetic data by DAFL, DeepInv, and sampled by DFND. (b) indicates the proportion of sample numbers in various classes (\%) of the original and substitute data.}
	\label{fig1}
	\vspace{-0.6cm}
\end{figure}

Since the original training data is not available for privacy or other reasons \cite{burton2015data}, the key is how to supplement the new training data, \textit{i.e.}, the substitution data.
Based on the source of the substitution data, almost all existing DFKD methods can be divided into generation-based and sampling-based methods.
Despite the impressive improvements achieved by these DFKD methods through complex loss stacking \cite{yin2020dreaming,fang2021contrastive} and knowledge distillation strategies \cite{fang2021mosaicking,fang2022up}, the trained students still suffer from distribution shifts between the substitution and original data, which has long been overlooked.
First, the quality of the synthetic or sampled images significantly differs from the original.
Besides, for generation-based methods, the synthetic data relies on the teacher's guidance, and it is easier to synthesize the class familiar to the generator.
For sampling-based methods, the sampled data entirely depends on the teacher's preference for various classes.
These protocols make the preference of the teacher model inevitably affect class proportions and also lead to distribution shifts.
Such shifts confound the student learning process.
For example, if a pre-trained teacher model is not familiar with a specific class $\mathcal{A}$, \textit{i.e.}, it is difficult to obtain high confidence, resulting in fewer synthetic or sampled data belonging to $\mathcal{A}$.
For the class balance, the teacher tends to classify ambiguous and indistinguishable data into $\mathcal{A}$, leading to the distribution shifts \cite{do2022momentum}.
Relying on these data, the student is inevitably confused with the original testing data with the different distributions.

More intrigued, we select three DFKD methods (DAFL \cite{chen2019data}, DeepInv \cite{yin2020dreaming}, and DFND \cite{chen2021learning}) and perform a toy experiment on the CIFAR-10 \cite{krizhevsky2009learning}.
This toy experiment aims to show the distribution shifts between the substitution and original data.
These methods include generation with generators (DAFL), generation through teacher model inversion (DeepInv), and sampling based on teacher preferences (DFND).
We use the original data as a comparison benchmark and compare them from two aspects: the quality of images and class proportions.
The results are shown in Figure~\ref{fig1}.
In Figure~\ref{fig1}\textcolor{red}{a}, we randomly visualize the original data, the substitution data of DAFL, DeepInv, and DFND, and calculate the Fr$\mathrm { \acute{e} }$chet Inception Distance (FID, lower is better) \cite{heusel2017gans}, a metric widely used to evaluate the quality of images.
The substitution and original data are different for the data distribution domain.
In Figure~\ref{fig1}\textcolor{red}{b}, we test the class proportions (the substitution data is based on teacher pseudo-labels).
A prominent result is that the classes of the substitution data are unbalanced due to teacher preferences, which greatly differ from the original data.
These observations confirm the distribution shifts between the substitution and original data, confounding the student model.

Based on these observations, we attempt to introduce a new perspective with causal inference to handle the distribution shifts.
During the application of theoretical causal inference \cite{pearl2009causal} to the DFKD task, the challenges lie in describing and designing plausible causal effects and identifying and compensating for biased student learning on the substitution data with shifts.
To this end, this paper attempts to address the challenges by drawing on instinctive human causalities~\cite{van1994cue} to find causal relationships among the variables in the DFKD task and optimize the biased student training process.
We first disentangle the causalities and customize the causal graph according to the properties of the variables in the DFKD task.
Based on this, we explore the causal paths from the substitution inputs $\bm{X}$ to the student predictions $\bm{S}$.
Then, we propose a simple yet effective Knowledge Distillation Causal Intervention (KDCI) framework to achieve de-confounded DFKD and use the do-calculus $P(\bm{S}|do(\bm{X}))$ to calculate the actual causal effect, instead of classic likelihood $P(\bm{S}|\bm{X})$ without considering the shifts.
KDCI can be easily combined with existing methods and use the backdoor adjustment \cite{glymour2016causal} to de-confound and alleviate the impact of the shifts.
Experiments on KDCI combined with six representative DFKD methods demonstrate its strong positive effect on the existing DFKD pipeline.
Specifically, the primary contributions and experiments are summarized below:
\begin{itemize}
    \item To our best knowledge, we are the first to alleviate the dilemma of the distribution shifts in the DFKD task from a causality-based perspective.
    Such shifts are regarded as the harmful confounder, which leads the student to learn misleading knowledge.
    \item We propose a KDCI framework to restrain the detrimental effect caused by the confounder and attempt to achieve the de-confounded distillation process.
    Besides, KDCI can be easily and flexibly combined with existing generation-based or sampling-based DFKD paradigms.
    \item Extensive experiments on the combination with six DFKD methods show that our KDCI can bring consistent and significant improvements to existing state-of-the-art models.
    Particularly, it improves the accuracy of the DeepInv~\cite{yin2020dreaming} by up to 15.54\% on the CIFAR-100 dataset.
\end{itemize}

\section{Related Work}
\textbf{Data-Free Knowledge Distillation.}
Data-free knowledge distillation is a promising task to train small models while avoiding leakage of original training data~\cite{lopes2017data,micaelli2019zero}.
The critical point is how to supplement substitution data~\cite{bhardwaj2019dream,luo2020large,choi2020data,wang2023explicit}.
The existing methods are mainly divided into three types: Generative Adversarial Networks (GANs) generation \cite{chen2019data,fang2022up}, teacher-based model inversion generation \cite{yin2020dreaming,fang2021contrastive}, and unlabeled data sampling \cite{fang2021mosaicking,chen2021learning,wang2023sampling}.
Chen \textit{et al.} \cite{chen2019data} introduce the generator into the DFKD task and improve teachers' familiarity with generating data.
Fang \textit{et al.} \cite{fang2022up} propose feature sharing to simplify the generation process.
To better generation quality, Yin \textit{et al.} \cite{yin2020dreaming} explore the prior knowledge of the data.
Fang \textit{et al.} \cite{fang2021contrastive} introduce contrastive learning to enhance student performance.
Chen \textit{et al.} \cite{chen2021learning} and Fang \textit{et al.} \cite{fang2021mosaicking} select wild data and out-of-domain (OOD) data to reduce generation costs.
Despite the promising performance, a long-overlooked issue is data distribution shifts, \textit{i.e.}, the distribution bias of the student's training data and the original data is a confounder that significantly causes performance bottlenecks.

\noindent \textbf{Causal Inference.}
Causal inference is a theory-oriented tool that seeks actual effects in a specific phenomenon~\cite{pearl2009causal}, which has been studied and followed by diverse fields such as economics~\cite{varian2016causal} and psychology~\cite{foster2010causal} communities. 
The mainstream causal inference studies applied to neural information processing consist of two aspects: intervention \cite{wang2020visual,chen2022causal,yang2021causal,liu2023learning,yang2024towards2,deng2021comprehensive} and counterfactuals \cite{tang2020unbiased,sun2022counterfactual,qian2021counterfactual,yang2024towards}.
Intervention is a technique for manipulating the original data distribution to reveal causal effects~\cite{glymour2016causal}.
Counterfactual describes the imagined results generated by factual variables when treated differently~\cite{pearl2009causality}.
Benefiting from the strong potential of causal inference to decouple spurious correlations among variables, it is gradually adopted to improve the performance of models for different downstream tasks, such as visual question answering~\cite{niu2021counterfactual}, emotion recognition~\cite{yang2023context}, and scene graph generation~\cite{tang2020unbiased}. In contrast, to our best knowledge, this is the first work to identify the distribution shifts in the DFKD task through the causal intervention and alleviate the confounding effect caused by the shifts.

\section{Methodology}

\subsection{Causal Graph of DFKD Task}

\begin{figure}[t]
	\centering
	\includegraphics[scale=0.33]{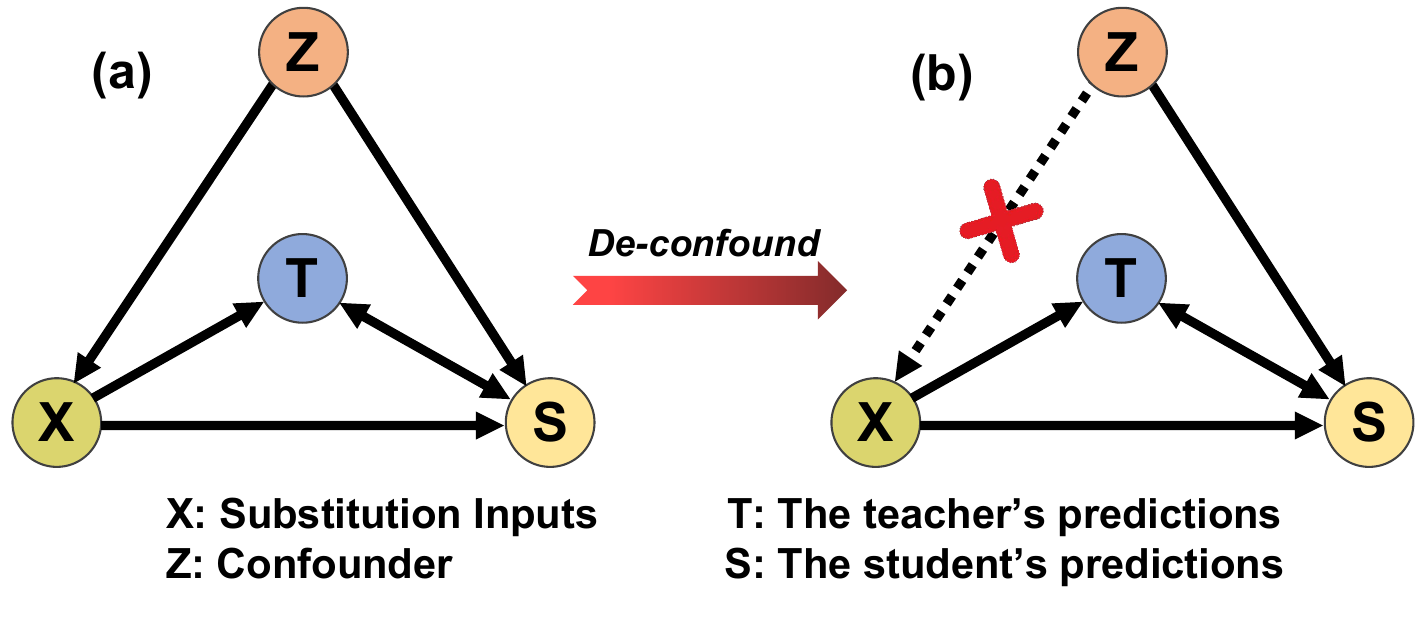}
	\vspace{-0.45cm}
	\caption{The causal graph. (a) The existing methods ignore distribution shifts. (b) The shifts are alleviated by causal inference.}
	\label{fig2}
    \vspace{-0.55cm}
\end{figure}

First, we customize the causal graph according to the properties of the variables in the DFKD task.
Specifically, the teacher is pre-trained with original training data, which is not disturbed by distribution shifts.
For the student, it uses the substitution data to train while testing on the original data.
The data distribution shifts indicate that it will be disturbed by the biased data \cite{do2022momentum}.
During the distillation process, the teacher and student are fed the same substitution data.
In this case, the student's predictions are constrained to learn the teacher's predictions.
Following the same graphical notation as \cite{pearl2000models} for clarity and interpretability, we denote the variables with the notes $\mathcal{N}$ and construct the direct causal effects with the links $\mathcal{E}$.
From Figure~\ref{fig2}, there are four variables involved in the DFKD causal graph $\mathcal{G} = \{ \mathcal{N}, \mathcal{E} \}$, which includes the substitution inputs $\bm{X}$, the confounder $\bm{Z}$, the teacher's predictions $\bm{T}$, and the student's predictions $\bm{S}$.
In particular, our causal graph is applicable to almost all existing DFKD methods so that it can be used as a general framework.
The details of the causal relationships are described as follows.

 $\bm{Z} \rightarrow \bm{X}$.
Existing DFKD methods rely on teacher predictions to supplement substitution data.
For the generation-based methods, the generator is guided by the teacher and more inclined to synthesize data that is easier to synthesize \cite{chen2019data,fang2022up,yin2020dreaming,fang2021contrastive}.
For the sampling-based methods, the data that the teacher is most \cite{chen2021learning} or least \cite{fang2021mosaicking} familiar with is sampled.
On the one hand, these synthetic or sampled data are always class-imbalanced.
On the other hand, these sources of substitution data rely heavily on the teacher, so they are highly volatile and vulnerable to teacher preferences.
These issues cause the distribution shifts between the original and substitution data.
The shifts are treated as the harmful confounder $\bm{Z}$~\cite{pearl2009causality}.
On this basis, the confounder $\bm{Z}$ causes the substitution data $\bm{X}$ to be biased compared to the original data, \textit{i.e.}, $\bm{Z}$ $\rightarrow$ $\bm{X}$.

$\bm{Z} \rightarrow \bm{S}$.
Due to the distribution shifts between the substitution and original data, the student trained on the substitution data tends to produce and exhibit biased predictions during the testing stage.
The detrimental confounder $\bm{Z}$ confounds and affects the student's training via the causal link $\bm{Z} \rightarrow \bm{S}$, which causes the performance bottleneck.

$\bm{X} \rightarrow \bm{T} / \bm{S}$ \& $\bm{T} \leftrightarrow \bm{S}$.
As with existing DFKD methods, both teacher and student make predictions on the substitution data $\bm{X}$ simultaneously.
By constraining their prediction distributions, the student's parameters are updated for optimization.
In our DFKD causal graph, the prediction processes of the teacher and student are represented as $\bm{X} \rightarrow \bm{T}$ and $\bm{X} \rightarrow \bm{S}$.
The link  $\bm{T} \leftrightarrow \bm{S}$ reflects the interaction causal effect between these two predictions during knowledge distillation.
Through these paths, the student can learn consistent knowledge from its teacher.

According to the causal theory \cite{pearl2009causal}, the confounder $\bm{Z}$ as a common cause directly or indirectly impacts the substitution inputs $\bm{X}$ and the student's predictions $\bm{S}$ simultaneously.
The knowledge transfer process from $\bm{T}$ to $\bm{S}$ increases the student's familiarity with these substitution data.
However, the confounder $\bm{Z}$ causes $\bm{X}$ to shift the original data distribution, leading to impure knowledge, which adversely affects student performance.
The detrimental effects follow the backdoor causal path as $\bm{X} \leftarrow \bm{Z} \rightarrow \bm{S}$.

\begin{figure*}[t]
	\centering
	\includegraphics[scale=0.4]{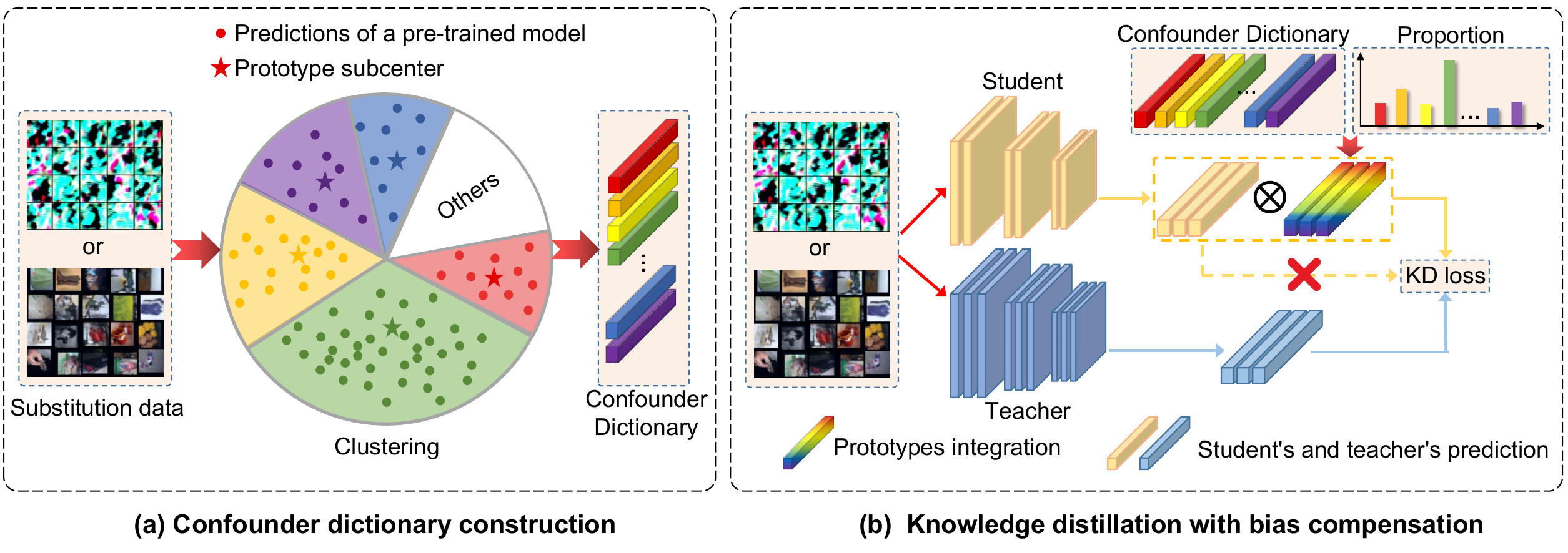}
    \vspace{-0.4cm}
	\caption{The overview of our KDCI. In stage (a), all substitution data is fed a pre-trained model to explore the prior knowledge and construct the confounder dictionary. 
    In stage (b), the prototype integration is built by the confounder dictionary and is used to compensate for biased student predictions. The distillation loss is calculated between the teacher's prediction and the student's compensated prediction.}
	\label{fig3}
    \vspace{-0.5cm}
\end{figure*}

\subsection{Causal Intervention via Backdoor Adjustment}

In the existing DFKD task, the pre-trained teacher model is fixed while the student model is learnable.
As shown in Figure~\ref{fig2}\textcolor{red}{a}, existing methods rely on the likelihood estimation of the student model as $P(\bm{S}|\bm{X})$.
The knowledge transfer process is expressed as:
\vspace{-0.1cm}
\begin{equation}\label{eq1}
\footnotesize
P(\bm{S}|\bm{X})\!=\!\sum_{\bm{z}}P(\bm{S}|\bm{X}\!, 
\!K\!D\!\left \langle\bm{T}\!=\!f_T(\bm{X}),\! \bm{S}\!=\!f_S(\bm{X},\bm{z})  \right \rangle )P(\bm{z}|\bm{X}),
\end{equation}

\noindent where $K\!D \left \langle \, , \right \rangle$ represents the knowledge distillation process between $\bm{T}$ and $\bm{S}$.
$f_T(\cdot)$ and $f_S(\cdot)$ represent the teacher model and the student model.
The confounder $\bm{Z}$ introduces the data distribution shifts via $P(\bm{z}|\bm{X})$, which makes the knowledge learned by the student impure.
To get rid of the confounding effect caused by $\bm{Z}$, an intuitive idea is changing inputs $\bm{X}$ to overcome the data distribution shifts and make $\bm{X}$ unaffected by $\bm{Z}$,
\textit{i.e.}, we have to use the data from the same distribution with the original training set as the student's training data.
However, it is not possible under the setting of the DFKD task.
To tackle this issue, we introduce the backdoor adjustment~\cite{pearl2009causal} to construct causal intervention $P(\bm{S}|do(\bm{X}))$ and block the backdoor path between $\bm{X}$ and $\bm{S}$ via $\bm{Z}$.
As a theoretical operation, implementing backdoor adjustment can be viewed as measuring the distribution shifts by estimating the average causal effect based on the class proportions.
By compensating for shifted student predictions, we alleviate the shift issue and suppress the disturbance of the confounder $\bm{Z}$.
In this case, the causal path from $\bm{Z}$ to $\bm{X}$ is cut-off in Figure~\ref{fig2}\textcolor{red}{b}.
The student learns pure knowledge with causal intervention $P(\bm{S}|do(\bm{X}))$ rather than original biased likelihood $P(\bm{S}|\bm{X})$.
This process can be expressed as:
\iffalse
\begin{equation}\label{eq2}
P_s(\bm{Y}_s|do(\bm{X}))\!\!=\!\!\sum_{\bm{z}}\!\!P_s(\bm{Y}_s|\bm{X}\!,\! K\!D\!\left \langle\bm{T}(\bm{X}),\! \bm{S}(\bm{X},\bm{z})\right \rangle)P_s(\bm{z}).
\end{equation}
\fi
\begin{equation}\label{eq2}
\footnotesize
P(\bm{S}|do(\bm{X}))\!=\!\sum_{\bm{z}}\!P(\bm{S}|\bm{X}\!,\! K\!D\!\left \langle\bm{T}\!=\!f_T(\bm{X}),\! \bm{S}\!=\!f_S(\bm{X},\bm{z})\right \rangle)P(\bm{z}),
\end{equation}
where $\bm{X}$ is no longer disturbed by $\bm{z}$ since causal intervention forces $\bm{X}$ to integrate each $\bm{z}$ fairly into the predictions of $\bm{S}$, according to the corresponding prior $P(\bm{z})$.

\subsection{De-confounded DFKD with KDCI}
To de-confound the DFKD task, we propose a Knowledge Distillation Causal Intervention (KDCI) framework to alleviate the distribution shift issue.
The overview of KDCI is shown in Figure~\ref{fig3}, which contains two stages: \textit{confounder dictionary construction} and \textit{knowledge distillation with bias compensation}.
First, after obtaining the substitution data and before training the student, we model the prior knowledge of these substitution data through the prototype clustering algorithm to obtain an intervention-driven confounder dictionary.
Then, the biased student predictions are compensated based on the subcenters and proportions.
Notably, for a general DFKD pipeline, our framework can be easily combined with other methods.
The implementation of KDCI is as follows.

\noindent \textbf{Confounder Dictionary Construction.}
Since the substitution data has no ground-truth information and the actual classes are ambiguous, we define a confounder dictionary $\bm{Z}=\left [ \bm{z}_1,\bm{z}_2, \dots, \bm{z}_{N} \right ]$ to explore the prior knowledge of these data.
$N$ is a hyperparameter representing the confounder size and $\bm{z}_i\in \mathbb{R}^d $ is a single prototype.
The prior knowledge implies the potential shifts and the differentiation information of class proportions.
From Figure~\ref{fig3}\textcolor{red}{a}, all substitution data is fed to an experienced pre-trained model (\textit{e.g.}, the teacher model itself) to obtain the prediction feature set $M = \left \{  m_j \in \mathbb{R}^d  \right \}_{j=1}^{N_m}$, where $N_m$ is the number of the substitution data.
We employ the K-Means++ with principle component analysis as the prototype clustering algorithm.
After clustering, each $\bm{z}_i$ represents a prototype feature cluster, and the prototype subcenter is put into the confounder dictionary as a prototype representation.
The feature cluster is denoted as $ {\textstyle \sum_{k=1}^{N_i}} m_k^i$
and the subcenter is denoted as $\bm{z}_i =  \frac{1}{N_i} {\textstyle \sum_{k=1}^{N_i}} m_k^i$, where $N_i$ is the number of the prediction features in $i$-th cluster.
Therefore, the prototype proportion can be calculated as $P(\bm{z}_i)=N_i/N_m$.

\noindent \textbf{Knowledge Distillation with Bias Compensation.}
After confounder dictionary construction, we approximate a theoretical causal inference by the confounder dictionary and prototype proportions to compensate for biased student predictions to learn pure knowledge, as shown in Figure~\ref{fig3}\textcolor{red}{b}.
In practice, the calculation of $P(\bm{S}|do(\bm{X}))$ requires multiple forward passes of all $\bm{z}$ resulting in expensive computational costs.
To simplify the above process, we apply the Normalized Weighted Geometric Mean (NWGM) \cite{xu2015show} and approximate the Eq.~(\ref{eq2}) as:
\iffalse
\begin{equation}\label{eq3}
P_s(\bm{Y}_s|do(\bm{X})){\approx} P_s(\bm{Y}_s|\bm{X}\!,\! K\!D \langle\bm{T}(\bm{X}),\! \sum_{\bm{z}}\!\!\bm{S}(\bm{X},\bm{z})P_s(\bm{z}) \rangle).
\end{equation}
\fi
\begin{equation}\label{eq3}
\footnotesize
P(\bm{S}|do(\bm{X})){\approx} P(\bm{S}|\bm{X}, K\!D \langle f_T(\bm{X}), \sum_{\bm{z}} f_S(\bm{X},\bm{z})P(\bm{z}) \rangle).
\end{equation}

\noindent During the knowledge transfer process, the update of student model parameters depends on the difference in predictions between the teacher and student, \textit{e.g.}, calculating the Kullback-Leibler (KL) divergence as $f_S \leftarrow \eta \nabla_s  K\!L (\bm{T}, \bm{S})$, where $\eta$ denotes learning rate, and $\nabla_s$ denotes the gradient.
Considering the distribution shift of training data, we introduce the prepared prior information of the cofounder dictionary to optimize the above process.
Based on this, the student predictions after compensation are represented as the integration of the biased predictions and the prior information as: $P(\bm{S}|do(\bm{X}))=\phi(f_S(\bm{X}), F(\bm{z}))$, where $\phi(\cdot)$ is a practically simple yet empirically powerful addition fusion strategy.
The prior information $F(\bm{z})$ is calculated as:

\vspace{-0.1cm}
\begin{equation}\label{eq4}
F(\bm{z})=\sum_{i=1}^{N}\lambda_i \bm{z}_i P(\bm{z}_i),
\end{equation}
where $\lambda_i$ is a weight coefficient that measures the importance of each prototype subcenter $\bm{z}_i$.
$P(\bm{z}_i)$ is the proportion of data in the \textit{i}-th cluster.
Here, we design an implementation of $\lambda_i$ with the additive attention as:
\begin{equation}\label{eq5}
\lambda_i = softmax(\bm{W}_t \cdot Tanh(\bm{W}_q f_S(\bm{X})+ \bm{W}_k \bm{z}_i)),
\end{equation}
where $\bm{W}_t \in \mathbb{R} ^{d_n\times 1}$, $\bm{W}_q \in \mathbb{R} ^{d_n\times d_h}$, and $\bm{W}_k \in \mathbb{R} ^{d_n\times d}$ are learnable mapping matrices.

\begin{table*}[t]
\centering
\caption{The accuracy (\%) on CIFAR-10 and CIFAR-100 about baseline methods vs. their KDCI-based version. \textbf{T.backbone} and \textbf{S.backbone} represent the backbones of the teacher and student.
\textbf{Teacher} and \textbf{Student} refer to scratch training on original data. The improved results are marked in \textbf{bold}.  $\{ \dagger, \ddagger, \natural, \sharp \}$ denote the provenance mentioned in the analysis.}
\vspace{-0.25cm}
\setlength{\tabcolsep}{3.2mm}
\scalebox{0.79}{
\begin{tabular}{@{}c|ccccc|ccccc@{}}
\toprule[1pt]
\textbf{Dataset}     & \multicolumn{5}{c|}{\textbf{CIFAR-10}}         & \multicolumn{5}{c}{\textbf{CIFAR-100}}         \\ \midrule
\textbf{T.backbone} & resnet-34 & vgg-11    & wrn-40-2 & wrn-40-2 & wrn-40-2 & resnet-34 & vgg-11    & wrn-40-2 & wrn-40-2 & wrn-40-2 \\
\textbf{S.backbone} & resnet-18 & resnet-18 & wrn-16-1 & wrn-40-1 & wrn-16-2 & resnet-18 & resnet-18 & wrn-16-1 & wrn-40-1 & wrn-16-2 \\ \midrule
\textbf{Teacher}           & 95.70 & 92.25 & 94.87 & 94.87 & 94.87 & 78.05 & 71.32 & 75.83 & 75.83 & 75.83 \\
\textbf{Student}           & 95.20 & 95.20 & 91.12 & 93.94 & 93.95 & 77.10 & 77.10 & 65.31 & 72.19 & 73.56 \\ \midrule
DAFL         & 92.22 & 81.10 & 65.71$^{\dagger}$ & 81.33 & 81.55 & 74.47 & 54.16 & 20.88$^{\sharp}$ & 42.83 & 43.70 \\
DAFL+\textbf{KDCI}    & \textbf{92.62} & \textbf{81.31} & \textbf{74.56}$^{\dagger}$ & \textbf{82.91} & \textbf{82.65} & \textbf{74.51} & \textbf{58.79} & \textbf{31.75}$^{\sharp}$ & \textbf{46.16} & \textbf{48.48} \\ \midrule
Fast         & 94.05 & 90.53 & 89.29 & 92.51 & 92.45 & 74.34 & 67.44 & 54.02 & 63.91 & 65.12 \\
Fast+\textbf{KDCI}    & \textbf{94.56} & \textbf{91.16} & \textbf{89.62} & \textbf{93.09} & \textbf{92.85} & \textbf{75.10} & \textbf{68.97} & \textbf{54.69} & \textbf{67.09} & \textbf{68.12} \\ \midrule
CMI          & 94.24 & 91.24 & 89.16 & 91.93 & 92.00 & 74.64 & 66.68 & 55.28 & 63.44 & 64.22 \\
CMI+\textbf{KDCI}     & \textbf{94.43} & \textbf{91.28} & \textbf{89.52} & \textbf{92.84} & \textbf{92.73} & \textbf{75.07} & \textbf{69.07} & \textbf{57.19} & \textbf{67.47} & \textbf{67.68} \\ \midrule
DeepInv      & 93.26 & 90.36 & 83.04 & 86.85 & 89.72 & 61.32$^{\natural}$ & 54.13$^{\ddagger}$ & 53.77 & 61.33 & 61.34 \\
DeepInv+\textbf{KDCI} & \textbf{93.67} & \textbf{91.42} & \textbf{83.47} & \textbf{89.32} & \textbf{91.06} & \textbf{74.59}$^{\natural}$ & \textbf{69.67}$^{\ddagger}$ & \textbf{55.22} & \textbf{62.13} & \textbf{65.90} \\ \midrule
Mosaick      & 95.27 & 91.69 & 90.03 & 93.28 & 92.94 & 75.91 & 71.58 & 59.32 & 66.61 & 67.36 \\
Mosaick+\textbf{KDCI} & \textbf{95.43} & \textbf{92.36} & \textbf{92.25} & \textbf{94.45} & \textbf{94.20} & \textbf{77.06} & \textbf{71.86} & \textbf{62.03} & \textbf{72.19} & \textbf{72.39} \\ \midrule
DFND         & 95.36 & 91.86 & 90.26 & 93.33 & 93.11 & 74.42 & 68.97 & 59.02 & 69.39 & 69.85 \\ 
DFND+\textbf{KDCI}    & \textbf{95.44} & \textbf{92.54} & \textbf{92.47} & \textbf{94.43} & \textbf{94.43} & \textbf{77.09} & \textbf{72.12} & \textbf{66.37} & \textbf{74.20} & \textbf{74.52} \\ \bottomrule[1pt]
\end{tabular}
}
\label{tab1}
\vspace{-0.5cm}
\end{table*}

\section{Experiments}
\subsection{Datasets and Models}
\textbf{Datasets.}
We evaluate the proposed framework on widely used classification datasets: CIFAR-10 \cite{krizhevsky2009learning}, CIFAR-100 \cite{krizhevsky2009learning}, Tiny-ImageNet \cite{le2015tiny}, and ImageNet \cite{deng2009imagenet}.
CIFAR-10 and CIFAR-100 contain 50,000 training samples and 10,000 testing samples of 32$\times$32 resolution.
Tiny-ImageNet contains 100,000 training samples, 10,000 validating samples, and 10,000 testing samples of 64$\times$64 resolution.
ImageNet contains 1000 classes with 1.28 million training samples and 50,000 validating samples of 224$\times$224 resolution.

\noindent \textbf{Models.}
We test the performance of various DFKD methods on several network architectures, including resnet \cite{he2016deep}, vgg \cite{simonyan2014very}, and wide resnet \cite{zagoruyko2016wide}.
For CIFAR-10  and  CIFAR-100, we use the pre-trained teacher models from CMI \cite{fang2021contrastive}, unify the teacher models among all methods, and set up five teacher-student backbone combinations following existing settings \cite{fang2021contrastive, fang2022up, fang2021mosaicking}.
For Tiny-ImageNet, we train a renset-34 teacher model without the mixup data augmentation \cite{zhang2017mixup}.
And the student utilizes the renset-18 as its backbone.
For ImageNet, we choose the same pre-trained resnet-50 teacher model with \cite{zhao2022decoupled} for all baseline methods.

\subsection{Method Zoo}
To comprehensively verify the effectiveness of KDCI, we select representative DFKD methods, including generation-based and sampling-based methods.
The generation-based methods spend extra computing costs to obtain substitute data by generative adversarial networks and teacher inversion, including DAFL \cite{chen2019data}, Fast \cite{fang2022up}, CMI \cite{fang2021contrastive}, and DeepInv \cite{bhardwaj2019dream}.
The sampling-based methods use unlabeled data as the substitute data, including Mosaick \cite{fang2021mosaicking} and DFND \cite{chen2021learning}.
For DAFL, Fast, and DeepInv, we follow the same settings as their original papers.
For CMI, due to the unpublished pre-inversion data, we choose the base version of CMI, which leads to the performance slightly lower than that reported in the original paper.
For Mosaick and DFND, we sample 600k unlabeled data in ImageNet \cite{deng2009imagenet} for CIFAR and Tiny-ImageNet, and 600k unlabeled data in Flicker1M dataset for ImageNet.
Due to the image quality, the reported performance of Mosaick is slightly better than the original paper.
The implementation details and loss functions of all the above methods are shown in \textbf{\textit{Supplementary Sec.7}}.

\begin{table}[t]
\centering
\caption{The accuracy (\%) on Tiny-ImageNet dataset. The teacher uses resnet-34, and the student uses resnet-18 as the backbones. The teacher achieves an accuracy of 52.74\%. The GPU time indicates the training time of one epoch on a single RTX 3090 GPU.}
\vspace{-0.25cm}
\setlength{\tabcolsep}{1.8mm}
\scalebox{0.78}{
\begin{tabular}{@{}c|ccc@{}}
\toprule
\textbf{Method}    & \textbf{Accuracy (\%)}                           & \textbf{GPU time}          & \textbf{Memory-Usage}          \\ \midrule
Fast      & 28.79                                   & 101.67s           & 5745M           \\
Fast+\textbf{KDCI} & \textbf{38.23} (+9.44) & \textbf{104.43s} (+2.71\%) & \textbf{5748M} (+0.05\%) \\ \midrule
DeepInv   & 20.68                                   & 255.26s           & 3312M           \\
DeepInv+\textbf{KDCI} & \textbf{34.84} (+14.16) & \textbf{258.51s} (+1.27\%) & \textbf{3316M} (+0.12\%) \\ \midrule
DFND      & 42.64                                   & 129.16s           & 4196M           \\
DFND+\textbf{KDCI} & \textbf{49.54} (+6.90) & \textbf{133.42s} (+3.30\%) & \textbf{4198M} (+0.05\%) \\ \bottomrule
\end{tabular}
}
\label{tab2}
\vspace{-0.55cm}
\end{table}

\subsection{Confounder Setup}
We use a pre-trained model to obtain the prediction feature set $M$.
By default, the pre-trained model is the teacher itself, which is trained on original data.
Each prediction feature $m$ is extracted from the logits output of the last layer, and the hidden dimension $d$ is equal to the number of classes.
By default, the number of clusters $N$ is the same across different datasets.
For the substitution data in a mini-batch of model inversion \cite{fang2021contrastive,bhardwaj2019dream}, the number of clusters $N$ is 32.
For the synthetic mini-batch from GANs \cite{chen2019data,fang2022up}, the number of clusters $N$ is 8.
For the unlabeled substitution data in sampling methods \cite{fang2021mosaicking,chen2021learning}, the number of clusters $N$ is 128.
Due to different training paradigms, the way KDCI is combined with these methods is different.
For the generation-based process, the generator and student models are updated alternately. 
We use a mini-batch of synthetic training data to construct the cofounder dictionary, and the dictionary will be updated as the generator is updated.
For the sampling-based process, unlabeled data only needs to be filtered once. 
We build the confounder dictionary once before distillation.
Pseudocode for the above processes and other training settings are shown in \textbf{\textit{Supplementary Sec.1}}.

\subsection{Performance Comparison}

To verify the proposed KDCI framework, we compare the original version and their KDCI-based version.

\noindent \textbf{Results on CIFAR-10 and CIFAR-100.} 
The results in Table~\ref{tab1} show the following vital observations.
(\textbf{\rmnum{1}})
KDCI consistently improves the performance of existing methods on all baselines across two datasets.
(\textbf{\rmnum{2}})
For CIFAR-10, although the original students' performance is already close to their teachers', KDCI still provides promising gains (mostly 1\%-2\% improvement) for students by eliminating the harmful impact of confounder.
For some baselines with poor results, KDCI brings significant improvement, \textit{e.g.}, up to 8.85\%$^{\dagger}$ for DAFL.
(\textbf{\rmnum{3}})
For CIFAR-100, KDCI can significantly improve various SOTA methods (about 3\%-5\% improvement on average).
Under some settings, KDCI improves the original methods with slightly lower performance to competitive performance, \textit{e.g.}, 15.54\%$^{\ddagger}$ and 13.27\%$^{\natural}$ for DeepInv \& 10.87\%$^{\sharp}$ for DAFL.
These strong gains demonstrate that KDCI can compensate for biased student predictions to learn pure knowledge by constructing prior knowledge on the substitution data whose data distribution differs from the original data distribution.
(\textbf{\rmnum{4}})
We notice a small increase for KDCI-based Fast \& CMI. 
The reasonable explanation is that they extract prior knowledge about the substitution data by accessing the statistics in the teacher's Batch Normalization layers \cite{ioffe2015batch}, which implicitly apply the likelihood estimation and weaken our causal intervention.
(\textbf{\rmnum{5}})
Besides, we are pleasantly surprised to find that the students trained by sampling-based methods (\textit{e.g.}, Mosaick \& DFND) can slightly outperform the teacher in some settings (\textit{e.g.}, vgg-11$\rightarrow $resnet-18), both the original and KDCI-based versions.
Both Mosaick and DFND utilize the unlabeled data.
With the additional rich semantic knowledge, more students outperform their teachers with the help of KDCI framework.

\begin{table}[t]
\centering
\caption{The accuracy (\%) on ImageNet dataset. ``$\rightarrow$'' denotes the teacher's (left) and student's (right) backbone pair.}
\vspace{-0.3cm}
\setlength{\tabcolsep}{2.2mm}
\scalebox{0.78}{
\begin{tabular}{@{}c|c|c@{}}
\toprule
\textbf{Settings}            & resnet-50 $\rightarrow $ resnet-18 & resnet-50 $\rightarrow $ mobilenetv2 \\ \midrule
Fast                & 53.45                              & 43.02                                \\
Fast+\textbf{KDCI}          & \textbf{58.24} (+4.79)                      & \textbf{50.12} (+7.10)                        \\ \midrule
Deeplnv             & 51.36                              & 40.25                                \\
Deeplnv+\textbf{KDCI}        &    \textbf{55.27} (+3.91)                                &       \textbf{46.24} (+5.99)                              \\ \midrule
DFND     & 42.82                              & 16.03                                \\
DFND+\textbf{KDCI} & \textbf{51.26} (+8.44)                      & \textbf{34.32} (+18.29)                       \\ 
\bottomrule
\end{tabular}
}
\label{tab}
\vspace{-0.55cm}
\end{table}

\noindent \textbf{Results on Tiny-ImageNet.} 
For the Tiny-ImageNet, we conduct experiments with Fast, DeepInv, and DFND.
The results are shown in Table~\ref{tab2}.
With the help of KDCI, the accuracy of the three methods is increased by 9.44\%, 14.16\%, and 6.90\%, respectively.
The Tiny-ImageNet dataset contains richer semantic information, which helps construct more expressive confounders and facilitates KDCI to bring more sufficient gains.
Besides, we test and show the additional calculation and memory overhead.
The overhead introduced by KDCI mainly comes from the confounder matrix. 
The additional overhead can be almost negligible since only a simple clustering algorithm is used.

\begin{table*}[t]
\centering
\caption{Ablation studies about the prior information $F(\bm{z})=\sum_{i=1}^{N}\lambda_i \bm{z}_i P(\bm{z}_i)$ in Eq.~(\ref{eq4}). The results include (1) original $F(\bm{z})$, (2) random weight coefficient $\lambda_i$, (3) random confounder dictionary $\bm{z}_i$, and (4) without (w/o) prototype proportion $P(\bm{z}_i)$.}
\vspace{-0.25cm}
\setlength{\tabcolsep}{3mm}
\scalebox{0.8}{
\begin{tabular}{@{}c|ccc|ccc|ccc|ccc@{}}
\toprule[1pt]
\textbf{Settings} &
  \multicolumn{3}{c|}{(1) \textbf{ Original $F(\bm{z})$}} &
  \multicolumn{3}{c|}{(2) \textbf{ Random $\lambda_i$}} &
  \multicolumn{3}{c|}{(3) \textbf{ Random $\bm{z_i}$}} &
  \multicolumn{3}{c}{(4) \textbf{w/o $P(\bm{z_i})$}} \\ \midrule
\textbf{Methods}   & Fast  & DeepInv & DFND  & Fast  & DeepInv & DFND  & Fast  & DeepInv & DFND  & Fast  & DeepInv & DFND  \\
\textbf{CIFAR-10}  & \textbf{94.56} & \textbf{93.67}   & \textbf{95.38} & 93.92 & 91.56   & 95.28 & 93.35 & 91.84   & 94.94 & 93.70 & 92.76   & 95.11 \\
\textbf{CIFAR-100} & \textbf{75.10} & \textbf{74.59}   & \textbf{77.09} & 74.79 & 72.72   & 76.86 & 73.76 & 72.81   & 76.14 & 74.60 & 72.66   & 76.97 \\ \bottomrule[1pt]
\end{tabular}
}
\label{tab3}
\vspace{-0.35cm}
\end{table*}

\begin{figure*}[h]
	\centering
	\includegraphics[scale=0.39]{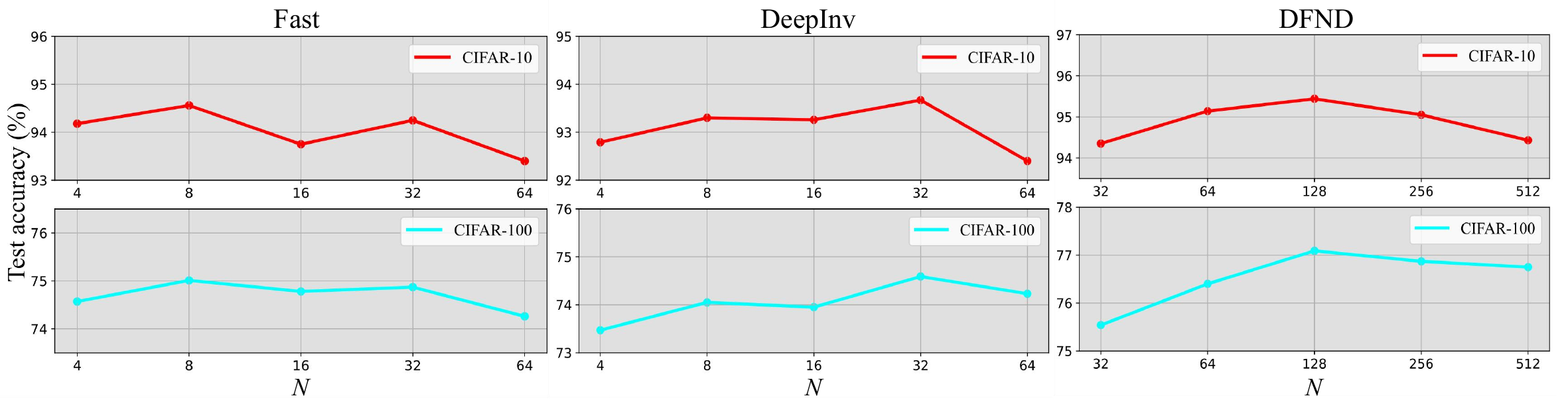}
	\vspace{-0.35cm}
	\caption{The test accuracy (\%) on CIFAR-10 and CIFAR-100 datasets about different confounder dictionary size $N$. The teacher uses resnet-34, and the student uses resnet-18 as the backbones.}
	\label{fig5}
	\vspace{-0.5cm}
\end{figure*}

\iffalse
Combining the performance of the above datasets, we conclude that KDCI can provide more significant help on CIFAR-100, Tiny-ImageNet \& ImageNet.
More complex datasets are more susceptible to teacher preferences, leading to more severe distribution shifts.
It is worth noting that for ImageNet, the mainstream methods update a thousand generators, which may somewhat alleviate distribution shifts. 
Nevertheless, KDCI still de-confound biased student predictions, achieving significant performance improvements.

\fi

\begin{table*}[t]
\centering
\caption{Ablation studies about the confounder dictionary $\bm{Z}$. 
``w/o $\bm{Z}$'' denotes the vanilla version of DFKD methods. ``\textit{original} $\bm{Z}$'' denotes the original confounder from the teacher itself. ``\textit{other} $\bm{Z}$'' denotes the confounder from another pre-trained model, \textit{i.e.}, 
swapping the confounder from the pre-training teacher models on CIFAR-10 and CIFAR-100 datasets.}
\vspace{-0.25cm}
\setlength{\tabcolsep}{3mm}
\scalebox{0.75}{
\begin{tabular}{@{}c|cccccc|cccccc@{}}
\toprule
\textbf{Dataset} &
  \multicolumn{6}{c|}{\textbf{CIFAR-10}} &
  \multicolumn{6}{c}{\textbf{CIFAR-100}} \\ \midrule
\textbf{Settings} &
  \multicolumn{3}{c|}{resnet-34 $\rightarrow $ resnet-18} &
  \multicolumn{3}{c|}{vgg-11 $\rightarrow $ resnet-18} &
  \multicolumn{3}{c|}{resnet-34 $\rightarrow $ resnet-18} &
  \multicolumn{3}{c}{vgg-11 $\rightarrow $ resnet-18} \\ \midrule
 $\bm{Z}$ &
  w/o $\bm{Z}$ &
  \textit{original} $\bm{Z}$ &
  \multicolumn{1}{c|}{\textit{other} $\bm{Z}$} &
  w/o $\bm{Z}$ &
  \textit{original} $\bm{Z}$ &
  \textit{other} $\bm{Z}$ &
  w/o $\bm{Z}$ &
  \textit{original} $\bm{Z}$ &
  \multicolumn{1}{c|}{\textit{other} $\bm{Z}$} &
  w/o $\bm{Z}$ &
  \textit{original} $\bm{Z}$ &
  \textit{other} $\bm{Z}$ \\ \midrule
\textbf{Fast} &
  94.05 &
  94.56 &
  \multicolumn{1}{c|}{93.96} &
  90.53 &
  91.16 &
  90.73 &
  74.42 &
  75.10 &
  \multicolumn{1}{c|}{74.75} &
  67.44 &
  68.97 &
  68.75 \\
\textbf{DeepInv} &
  93.26 &
  93.67 &
  \multicolumn{1}{c|}{93.56} &
  90.36 &
  91.42 &
  91.26 &
  61.32 &
  74.59 &
  \multicolumn{1}{c|}{73.04} &
  54.13 &
  69.67 &
  68.04 \\
\textbf{DFND} &
  95.36 &
  95.44 &
  \multicolumn{1}{c|}{95.41} &
  91.86 &
  92.54 &
  92.34 &
  74.34 &
  77.09 &
  \multicolumn{1}{c|}{76.97} &
  68.97 &
  72.12 &
  71.97 \\ \bottomrule
\end{tabular}
}
\label{tab4}
\vspace{-0.15cm}
\end{table*}

\noindent \textbf{Results on ImageNet.} 
For the ImageNet, we conduct two backbone combinations with three baseline methods.
The results are shown in Table~\ref{tab}.
The generation-based methods (Fast \& Deeplnv) have to train 1,000 generators (one generator for one class).
We speculate that a possible reason why KDCI has smaller gains for these two generation-based methods is that ‘one generator for one class’ may alleviate the distribution shifts issue to a certain extent and thereby weaken the effect of causal intervention. 
In comparison, the gain of KDCI for DFND is higher.
Overall, from the experimental results of ImageNet, the positive impact of KDCI on students is also consistent.
These results further validate the effectiveness of our method.

Combining the performance on the above datasets, we conclude that KDCI can provide more significant help on more complex datasets (\textit{e.g.}, ImageNet \& Tiny-ImageNet with more classes and various visual effects).
More complex datasets are more susceptible to teacher preferences, leading to more severe distribution shifts.
Further, the detrimental shifts inevitably lead to biased substitution data compared to the original data.
Fortunately, KDCI favorably de-confound the biased student predictions, achieving significant performance improvements.

\begin{figure*}[t]
	\centering
    \vspace{-0.1cm}
	\includegraphics[scale=0.46]{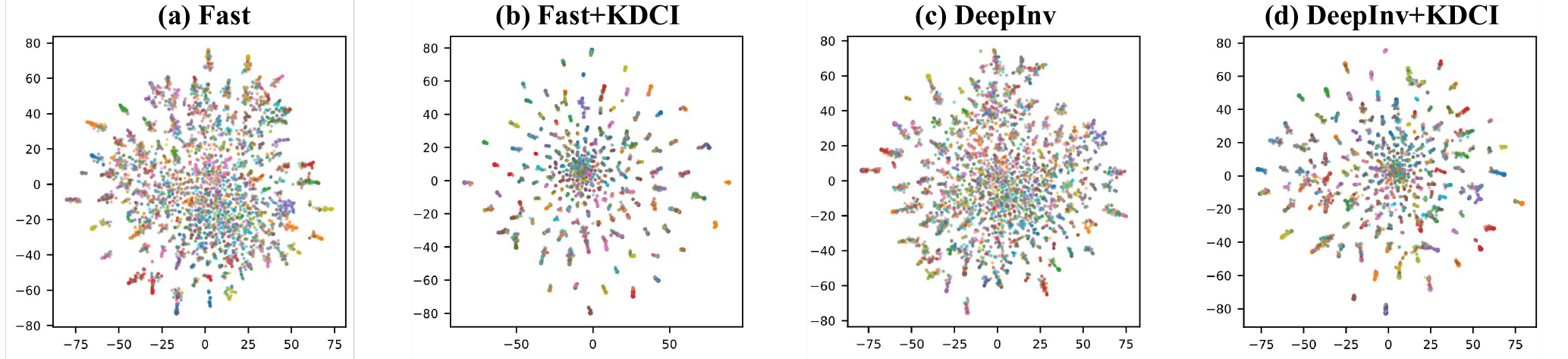}
	\vspace{-0.3cm}
	\caption{T-SNE results of vanilla and KDCI-based models performance on Tiny-ImageNet dataset. KDCI helps models obtain clearer clustering results, which show its strong positive impact.}
	\label{fig6}
	\vspace{-0.45cm}
\end{figure*}

\subsection{Analysis of Prior Information $F(\bm{z})$}

We conduct ablation studies to validate the effectiveness of the components of prior information $F(\bm{z})$ in Eq.~(\ref{eq4}) used to compensate students for biased predictions in Table~\ref{tab3}.
We select three methods (Fast, DeepInv, and DFND) on both CIFAR-10 and CIFAR-100 datasets.
The teacher and student use resnet-34 and resnet-18 as their backbones, respectively.
Other settings are the same as Table~\ref{tab1}.

\noindent \textbf{Necessity of Weight Coefficient $\lambda_i$.}
The weight $\lambda_i$ represents the degree of each confounder. 
Comparing (1) and (2), the random $\lambda_i$ causes a decline in performance.
Such results indicate that depicting the importance of each confounder is essential to achieve effective causal intervention.

\noindent \textbf{Rationality of Confounder $\bm{z}_i$.}
The confounder $\bm{z}_i$ comes from the predicted feature representation of the pre-trained model, which directly implies prior knowledge about the substitution data.
Comparing (1) and (3), students using our custom confounder significantly outperform the alternative confounder that are randomly initialized, which proves the validity of extracted prior knowledge.

\noindent \textbf{Impact of Prototype Proportion $P(\bm{z}_i)$.}
The prototype proportion $P(\bm{z}_i)$ denotes the frequency of each confounder containing the knowledge of feature proportions.
From (1) and (4), the proportion of each confounder plays a vital role in precise intervention implementation.

\subsection{Analysis of Confounder Dictionary $\bm{Z}$}

The confounder dictionary $\bm{Z}$ is proposed to explore the prior knowledge of the substitution data.
We investigate the effectiveness of $\bm{Z}$ in two perspectives: the confounder prototype size $N$ and the selected pre-trained models.
For the size, we select representative methods to test the effect of different $N$.
For the selected pre-trained models, we use the models coming from other datasets with different numbers of classes.
We swap the pre-training models on CIFAR-10 and CIFAR-100 to build the confounder and align the feature dimensions through a learnable mapping matrix.

\noindent \textbf{Impact of Confounder Dictionary Size $N$.}
To justify the size $N$ of the confounder $\bm{Z}$, we set five sets of $N$ for each method.
For Fast and DeepInv, $\bm{Z}$ comes from a mini-batch synthetic data.
For DFND, $\bm{Z}$ comes from the sampled data.
In Figure~\ref{fig5}, designing the suitable $N$ for methods that suffer from varying degrees of harmful shifts helps to perform de-confounded training better.

\noindent \textbf{Impact of Confounder Dictionary Sources.}
Table~\ref{tab4} shows three settings with/without confounder dictionary $\bm{Z}$.
We have two interesting discoveries.
(\textbf{\rmnum{1}}) First, an obvious conclusion is that using $\bm{Z}$ outperforms the original DFKD methods without $\bm{Z}$ in almost all settings.
Such observations demonstrate the effectiveness of causal intervention.
(\textbf{\rmnum{2}}) Second, swapping the confounders from CIFAR-10 and CIFAR-100 teacher models brings the performance decrease.
For CIFAR-10, the distribution of the substitution data is simple.
Simple distributions are over-separated when features are extracted using pre-trained models from complex distributions.
We call this phenomenon \textit{over-intervention}.
The excessive causal intervention potentially causes the deviation of the confounder itself.
For CIFAR-100, the distribution is more complex.
The complex distributions are not well approximated when using pre-trained models with less discriminative ability.
We call this phenomenon \textit{under-intervention}.
The incomplete causal intervention would lead to gain reduction.

\begin{figure}[t]
	\centering
    \vspace{-0.1cm}
	\includegraphics[scale=0.56]{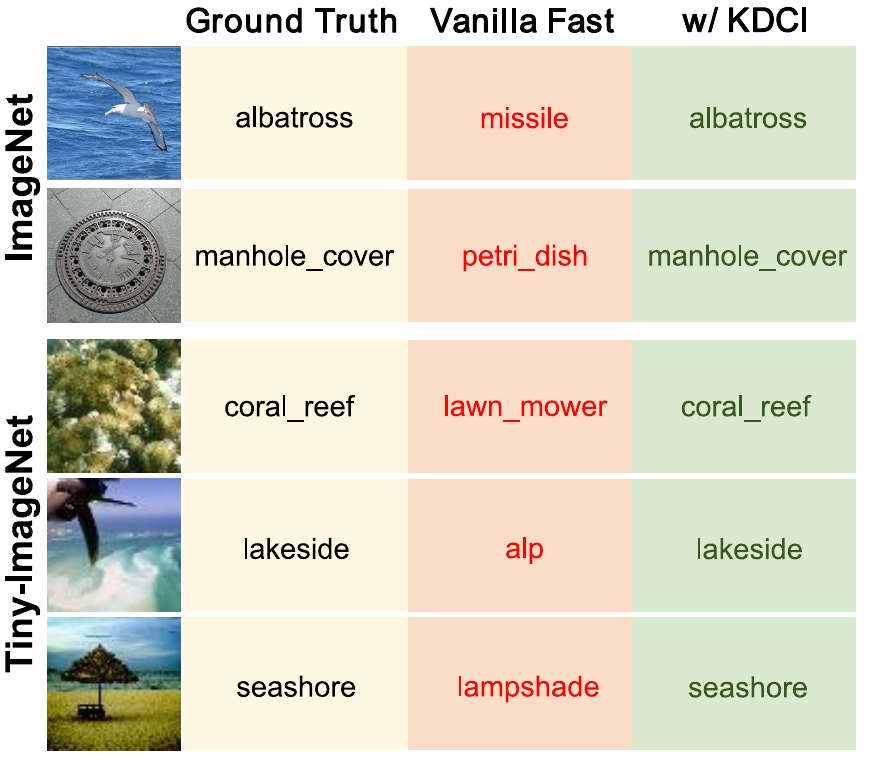}
	\vspace{-0.8cm}
	\caption{Qualitative results of the vanilla and KDCI-based version on ImageNet and Tiny-ImageNet.}
	\label{fig}
	\vspace{-0.55cm}
\end{figure}

\subsection{Qualitative Results}

Further, we present qualitative results to further demonstrate the positive gains of KDCI over baseline methods.

\noindent \textbf{Visualization Results.}
To intuitively show the help of KDCI to existing DFKD methods, we first visualize the student classification results with t-SNE \cite{van2008visualizing} on the Tiny-ImageNet dataset.
We reserve 100 classes of validating samples.
From Figure~\ref{fig6}, the KDCI-based versions (b)\&(d) have fewer outliers and clearer clustering effects than the vanilla versions (a)\&(c).
These phenomena further confirm that our KDCI can well disentangle features from different classes, thus improving existing methods' performance.

\noindent \textbf{Case Study of Causal Intervention.}
We select representative examples from ImageNet and Tiny-ImageNet datasets to show differences in student predictions before and after the intervention.
As shown in Figure~\ref{fig}, KDCI can eliminate the prediction offset caused by some misleading features to a certain extent.
For example, students from the vanilla Fast misclassify ``albatross" as ``missile" or ``coral\_reef'' as ``lawn\_mower'' due to large patches of similar background colour, 
and misclassify ``manhole\_cover" as ``petri\_dish" or ``seashore'' as ``lampshade'' due to similar shape. 
Fortunately, KDCI can repair prediction shifts in the above cases.

\section{Conclusion}
This paper proposes a novel perspective from causal inference to handle the distribution shifts in the Data-Free Knowledge Distillation (DFKD) task.
By customizing the causal graph according to the properties of the variables in the DFKD, we propose a Knowledge Distillation Causal Intervention (KDCI) framework to de-confound the adverse effect caused by the shifts between the substitution and original data.
KDCI can be flexibly combined with most existing methods.
Numerous experiments prove that KDCI can consistently help existing methods and provide an alternative causal intervention insight.

\section*{Acknowledgements}
%This work is supported by Shanghai Municipal Science and Technology Major Project (No.2021SHZDZX0103), the Shanghai Engineering Research Center of AI \& Robotics, Fudan University, China, the Engineering Research Center of AI \& Robotics, Ministry of Education, China and the Green Ecological Smart Technology School-Enterprise Joint Research Center.
This work is supported by the Shanghai Engineering Research Center of AI \& Robotics, Fudan University, China, the Engineering Research Center of AI \& Robotics, Ministry of Education, China, and the Green Ecological Smart Technology School-Enterprise Joint Research Center.

\newpage
{
    \small
    \bibliographystyle{unsrt}
    \bibliography{arxiv}

\begin{thebibliography}{10}

\bibitem{he2016deep}
Kaiming He, Xiangyu Zhang, Shaoqing Ren, and Jian Sun.
\newblock Deep residual learning for image recognition.
\newblock In {\em Proceedings of the IEEE/CVF Conference on Computer Vision and Pattern Recognition (CVPR)}, pages 770--778, 2016.

\bibitem{liu2021swin}
Ze~Liu, Yutong Lin, Yue Cao, Han Hu, Yixuan Wei, Zheng Zhang, Stephen Lin, and Baining Guo.
\newblock Swin transformer: Hierarchical vision transformer using shifted windows.
\newblock In {\em Proceedings of the IEEE/CVF International Conference on Computer Vision (ICCV)}, pages 10012--10022, 2021.

\bibitem{yang2023target}
Dingkang Yang, Yang Liu, Can Huang, Mingcheng Li, Xiao Zhao, Yuzheng Wang, Kun Yang, Yan Wang, Peng Zhai, and Lihua Zhang.
\newblock Target and source modality co-reinforcement for emotion understanding from asynchronous multimodal sequences.
\newblock {\em Knowledge-Based Systems}, page 110370, 2023.

\bibitem{liu2023stochastic}
Yang Liu, Dingkang Yang, Gaoyun Fang, Yuzheng Wang, Donglai Wei, Mengyang Zhao, Kai Cheng, Jing Liu, and Liang Song.
\newblock Stochastic video normality network for abnormal event detection in surveillance videos.
\newblock {\em Knowledge-Based Systems}, 280:110986, 2023.

\bibitem{yang2024robust}
Dingkang Yang, Kun Yang, Mingcheng Li, Shunli Wang, Shuaibing Wang, and Lihua Zhang.
\newblock Robust emotion recognition in context debiasing.
\newblock {\em arXiv preprint arXiv:2403.05963}, 2024.

\bibitem{yang2022disentangled}
Dingkang Yang, Shuai Huang, Haopeng Kuang, Yangtao Du, and Lihua Zhang.
\newblock Disentangled representation learning for multimodal emotion recognition.
\newblock In {\em Proceedings of the 30th ACM International Conference on Multimedia (ACM MM)}, pages 1642--1651, 2022.

\bibitem{yang2022learning}
Dingkang Yang, Haopeng Kuang, Shuai Huang, and Lihua Zhang.
\newblock Learning modality-specific and -agnostic representations for asynchronous multimodal language sequences.
\newblock In {\em Proceedings of the 30th ACM International Conference on Multimedia (ACM MM)}, pages 1708--1717, 2022.

\bibitem{devlin2018bert}
Jacob Devlin, Ming-Wei Chang, Kenton Lee, and Kristina Toutanova.
\newblock Bert: Pre-training of deep bidirectional transformers for language understanding.
\newblock {\em arXiv preprint arXiv:1810.04805}, 2018.

\bibitem{karras2019style}
Tero Karras, Samuli Laine, and Timo Aila.
\newblock A style-based generator architecture for generative adversarial networks.
\newblock In {\em Proceedings of the IEEE/CVF Conference on Computer Vision and Pattern Recognition (CVPR)}, pages 4401--4410, 2019.

\bibitem{brown2020language}
Tom Brown, Benjamin Mann, Nick Ryder, Melanie Subbiah, Jared~D Kaplan, Prafulla Dhariwal, Arvind Neelakantan, Pranav Shyam, Girish Sastry, Amanda Askell, et~al.
\newblock Language models are few-shot learners.
\newblock {\em Advances in Neural Information Processing Systems (NeurIPS)}, 33:1877--1901, 2020.

\bibitem{yang2024how2comm}
Dingkang Yang, Kun Yang, Yuzheng Wang, Jing Liu, Zhi Xu, Rongbin Yin, Peng Zhai, and Lihua Zhang.
\newblock How2comm: Communication-efficient and collaboration-pragmatic multi-agent perception.
\newblock {\em Advances in Neural Information Processing Systems (NeurIPS)}, 36, 2024.

\bibitem{chen2022towards}
Zhaoyu Chen, Bo~Li, Jianghe Xu, Shuang Wu, Shouhong Ding, and Wenqiang Zhang.
\newblock Towards practical certifiable patch defense with vision transformer.
\newblock In {\em Proceedings of the IEEE/CVF Conference on Computer Vision and Pattern Recognition (CVPR)}, pages 15148--15158, 2022.

\bibitem{liu2023amp}
Yang Liu, Jing Liu, Kun Yang, Bobo Ju, Siao Liu, Yuzheng Wang, Dingkang Yang, Peng Sun, and Liang Song.
\newblock Amp-net: Appearance-motion prototype network assisted automatic video anomaly detection system.
\newblock {\em IEEE Transactions on Industrial Informatics}, 2023.

\bibitem{radford2021learning}
Alec Radford, Jong~Wook Kim, Chris Hallacy, Aditya Ramesh, Gabriel Goh, Sandhini Agarwal, Girish Sastry, Amanda Askell, Pamela Mishkin, Jack Clark, et~al.
\newblock Learning transferable visual models from natural language supervision.
\newblock In {\em International Conference on Machine Learning (ICML)}, pages 8748--8763. PMLR, 2021.

\bibitem{ramesh2022hierarchical}
Aditya Ramesh, Prafulla Dhariwal, Alex Nichol, Casey Chu, and Mark Chen.
\newblock Hierarchical text-conditional image generation with clip latents.
\newblock {\em arXiv preprint arXiv:2204.06125}, 2022.

\bibitem{deng2009imagenet}
Jia Deng, Wei Dong, Richard Socher, Li-Jia Li, Kai Li, and Li~Fei-Fei.
\newblock Imagenet: A large-scale hierarchical image database.
\newblock In {\em Proceedings of the IEEE/CVF Conference on Computer Vision and Pattern Recognition (CVPR)}, pages 248--255. Ieee, 2009.

\bibitem{yang2023aide}
Dingkang Yang, Shuai Huang, Zhi Xu, Zhenpeng Li, Shunli Wang, Mingcheng Li, Yuzheng Wang, Yang Liu, Kun Yang, Zhaoyu Chen, et~al.
\newblock Aide: A vision-driven multi-view, multi-modal, multi-tasking dataset for assistive driving perception.
\newblock In {\em Proceedings of the IEEE/CVF International Conference on Computer Vision}, pages 20459--20470, 2023.

\bibitem{wang2023adversarial}
Yuzheng Wang, Zhaoyu Chen, Dingkang Yang, Yang Liu, Siao Liu, Wenqiang Zhang, and Lizhe Qi.
\newblock Adversarial contrastive distillation with adaptive denoising.
\newblock In {\em ICASSP 2023-2023 IEEE International Conference on Acoustics, Speech and Signal Processing (ICASSP)}, pages 1--5. IEEE, 2023.

\bibitem{ge2022zoom}
Zuhao Ge, Lizhe Qi, Yuzheng Wang, and Yunquan Sun.
\newblock Zoom-and-reasoning: Joint foreground zoom and visual-semantic reasoning detection network for aerial images.
\newblock {\em IEEE Signal Processing Letters}, 29:2572--2576, 2022.

\bibitem{liu2023improving}
Siao Liu, Zhaoyu Chen, Yang Liu, Yuzheng Wang, Dingkang Yang, Zhile Zhao, Ziqing Zhou, Xie Yi, Wei Li, Wenqiang Zhang, et~al.
\newblock Improving generalization in visual reinforcement learning via conflict-aware gradient agreement augmentation.
\newblock In {\em Proceedings of the IEEE/CVF International Conference on Computer Vision (ICCV)}, pages 23436--23446, 2023.

\bibitem{lopes2017data}
Raphael~Gontijo Lopes, Stefano Fenu, and Thad Starner.
\newblock Data-free knowledge distillation for deep neural networks.
\newblock {\em arXiv preprint arXiv:1710.07535}, 2017.

\bibitem{hinton2015distilling}
Geoffrey Hinton, Oriol Vinyals, Jeff Dean, et~al.
\newblock Distilling the knowledge in a neural network.
\newblock {\em arXiv preprint arXiv:1503.02531}, 2(7), 2015.

\bibitem{romero2014fitnets}
Adriana Romero, Nicolas Ballas, Samira~Ebrahimi Kahou, Antoine Chassang, Carlo Gatta, and Yoshua Bengio.
\newblock Fitnets: Hints for thin deep nets.
\newblock {\em arXiv preprint arXiv:1412.6550}, 2014.

\bibitem{wang2024out}
Yuzheng Wang, Zhaoyu Chen, Dingkang Yang, Pinxue Guo, Kaixun Jiang, Wenqiang Zhang, and Lizhe Qi.
\newblock Out of thin air: Exploring data-free adversarial robustness distillation.
\newblock In {\em Proceedings of the AAAI Conference on Artificial Intelligence}, volume~38, pages 5776--5784, 2024.

\bibitem{burton2015data}
Paul~R Burton, Madeleine~J Murtagh, Andy Boyd, James~B Williams, Edward~S Dove, Susan~E Wallace, Anne-Marie Tasse, Julian Little, Rex~L Chisholm, Amadou Gaye, et~al.
\newblock Data safe havens in health research and healthcare.
\newblock {\em Bioinformatics}, 31(20):3241--3248, 2015.

\bibitem{yin2020dreaming}
Hongxu Yin, Pavlo Molchanov, Jose~M Alvarez, Zhizhong Li, Arun Mallya, Derek Hoiem, Niraj~K Jha, and Jan Kautz.
\newblock Dreaming to distill: Data-free knowledge transfer via deepinversion.
\newblock In {\em Proceedings of the IEEE/CVF Conference on Computer Vision and Pattern Recognition (CVPR)}, pages 8715--8724, 2020.

\bibitem{fang2021contrastive}
Gongfan Fang, Jie Song, Xinchao Wang, Chengchao Shen, Xingen Wang, and Mingli Song.
\newblock Contrastive model inversion for data-free knowledge distillation.
\newblock {\em arXiv preprint arXiv:2105.08584}, 2021.

\bibitem{fang2021mosaicking}
Gongfan Fang, Yifan Bao, Jie Song, Xinchao Wang, Donglin Xie, Chengchao Shen, and Mingli Song.
\newblock Mosaicking to distill: Knowledge distillation from out-of-domain data.
\newblock {\em Advances in Neural Information Processing Systems (NeurIPS)}, 34:11920--11932, 2021.

\bibitem{fang2022up}
Gongfan Fang, Kanya Mo, Xinchao Wang, Jie Song, Shitao Bei, Haofei Zhang, and Mingli Song.
\newblock Up to 100x faster data-free knowledge distillation.
\newblock In {\em Proceedings of the AAAI Conference on Artificial Intelligence (AAAI)}, volume~36, pages 6597--6604, 2022.

\bibitem{do2022momentum}
Kien Do, Thai~Hung Le, Dung Nguyen, Dang Nguyen, Haripriya Harikumar, Truyen Tran, Santu Rana, and Svetha Venkatesh.
\newblock Momentum adversarial distillation: Handling large distribution shifts in data-free knowledge distillation.
\newblock {\em Advances in Neural Information Processing Systems (NeurIPS)}, 35:10055--10067, 2022.

\bibitem{chen2019data}
Hanting Chen, Yunhe Wang, Chang Xu, Zhaohui Yang, Chuanjian Liu, Boxin Shi, Chunjing Xu, Chao Xu, and Qi~Tian.
\newblock Data-free learning of student networks.
\newblock In {\em Proceedings of the IEEE/CVF International Conference on Computer Vision (ICCV)}, pages 3514--3522, 2019.

\bibitem{chen2021learning}
Hanting Chen, Tianyu Guo, Chang Xu, Wenshuo Li, Chunjing Xu, Chao Xu, and Yunhe Wang.
\newblock Learning student networks in the wild.
\newblock In {\em Proceedings of the IEEE/CVF Conference on Computer Vision and Pattern Recognition (CVPR)}, pages 6428--6437, 2021.

\bibitem{krizhevsky2009learning}
Alex Krizhevsky, Geoffrey Hinton, et~al.
\newblock Learning multiple layers of features from tiny images.
\newblock 2009.

\bibitem{heusel2017gans}
Martin Heusel, Hubert Ramsauer, Thomas Unterthiner, Bernhard Nessler, and Sepp Hochreiter.
\newblock Gans trained by a two time-scale update rule converge to a local nash equilibrium.
\newblock {\em Advances in Neural Information Processing Systems (NeurIPS)}, 30, 2017.

\bibitem{pearl2009causal}
Judea Pearl.
\newblock Causal inference in statistics: An overview.
\newblock {\em Statistics Surveys}, 3:96--146, 2009.

\bibitem{van1994cue}
Linda~J Van~Hamme and Edward~A Wasserman.
\newblock Cue competition in causality judgments: The role of nonpresentation of compound stimulus elements.
\newblock {\em Learning and Motivation}, 25(2):127--151, 1994.

\bibitem{glymour2016causal}
Madelyn Glymour, Judea Pearl, and Nicholas~P Jewell.
\newblock {\em Causal inference in statistics: A primer}.
\newblock John Wiley \& Sons, 2016.

\bibitem{micaelli2019zero}
Paul Micaelli and Amos~J Storkey.
\newblock Zero-shot knowledge transfer via adversarial belief matching.
\newblock {\em Advances in Neural Information Processing Systems (NeurIPS)}, 32, 2019.

\bibitem{bhardwaj2019dream}
Kartikeya Bhardwaj, Naveen Suda, and Radu Marculescu.
\newblock Dream distillation: A data-independent model compression framework.
\newblock {\em arXiv preprint arXiv:1905.07072}, 2019.

\bibitem{luo2020large}
Liangchen Luo, Mark Sandler, Zi~Lin, Andrey Zhmoginov, and Andrew Howard.
\newblock Large-scale generative data-free distillation.
\newblock {\em arXiv preprint arXiv:2012.05578}, 2020.

\bibitem{choi2020data}
Yoojin Choi, Jihwan Choi, Mostafa El-Khamy, and Jungwon Lee.
\newblock Data-free network quantization with adversarial knowledge distillation.
\newblock In {\em Proceedings of the IEEE/CVF Conference on Computer Vision and Pattern Recognition Workshops (CVPRW)}, pages 710--711, 2020.

\bibitem{wang2023explicit}
Yuzheng Wang, Zuhao Ge, Zhaoyu Chen, Xian Liu, Chuangjia Ma, Yunquan Sun, and Lizhe Qi.
\newblock Explicit and implicit knowledge distillation via unlabeled data.
\newblock In {\em ICASSP 2023-2023 IEEE International Conference on Acoustics, Speech and Signal Processing (ICASSP)}, pages 1--5. IEEE, 2023.

\bibitem{wang2023sampling}
Yuzheng Wang, Zhaoyu Chen, Jie Zhang, Dingkang Yang, Zuhao Ge, Yang Liu, Siao Liu, Yunquan Sun, Wenqiang Zhang, and Lizhe Qi.
\newblock Sampling to distill: Knowledge transfer from open-world data.
\newblock {\em arXiv preprint arXiv:2307.16601}, 2023.

\bibitem{varian2016causal}
Hal~R Varian.
\newblock Causal inference in economics and marketing.
\newblock {\em Proceedings of the National Academy of Sciences}, 113(27):7310--7315, 2016.

\bibitem{foster2010causal}
E~Michael Foster.
\newblock Causal inference and developmental psychology.
\newblock {\em Developmental Psychology}, 46(6):1454, 2010.

\bibitem{wang2020visual}
Tan Wang, Jianqiang Huang, Hanwang Zhang, and Qianru Sun.
\newblock Visual commonsense r-cnn.
\newblock In {\em Proceedings of the IEEE/CVF Conference on Computer Vision and Pattern Recognition (CVPR)}, pages 10760--10770, 2020.

\bibitem{chen2022causal}
Yingjie Chen, Diqi Chen, Tao Wang, Yizhou Wang, and Yun Liang.
\newblock Causal intervention for subject-deconfounded facial action unit recognition.
\newblock {\em arXiv preprint arXiv:2204.07935}, 2022.

\bibitem{yang2021causal}
Xu~Yang, Hanwang Zhang, Guojun Qi, and Jianfei Cai.
\newblock Causal attention for vision-language tasks.
\newblock In {\em Proceedings of the IEEE/CVF Conference on Computer Vision and Pattern Recognition (CVPR)}, pages 9847--9857, 2021.

\bibitem{liu2023learning}
Yang Liu, Zhaoyang Xia, Mengyang Zhao, Donglai Wei, Yuzheng Wang, Siao Liu, Bobo Ju, Gaoyun Fang, Jing Liu, and Liang Song.
\newblock Learning causality-inspired representation consistency for video anomaly detection.
\newblock In {\em Proceedings of the 31st ACM International Conference on Multimedia (ACM MM)}, pages 203--212, 2023.

\bibitem{yang2024towards2}
Dingkang Yang, Dongling Xiao, Ke~Li, Yuzheng Wang, Zhaoyu Chen, Jinjie Wei, and Lihua Zhang.
\newblock Towards multimodal human intention understanding debiasing via subject-deconfounding.
\newblock {\em arXiv preprint arXiv:2403.05025}, 2024.

\bibitem{deng2021comprehensive}
Xiang Deng and Zhongfei Zhang.
\newblock Comprehensive knowledge distillation with causal intervention.
\newblock {\em Advances in Neural Information Processing Systems (NeurIPS)}, 34:22158--22170, 2021.

\bibitem{tang2020unbiased}
Kaihua Tang, Yulei Niu, Jianqiang Huang, Jiaxin Shi, and Hanwang Zhang.
\newblock Unbiased scene graph generation from biased training.
\newblock In {\em Proceedings of the IEEE/CVF Conference on Computer Vision and Pattern Recognition (CVPR)}, pages 3716--3725, 2020.

\bibitem{sun2022counterfactual}
Teng Sun, Wenjie Wang, Liqiang Jing, Yiran Cui, Xuemeng Song, and Liqiang Nie.
\newblock Counterfactual reasoning for out-of-distribution multimodal sentiment analysis.
\newblock {\em arXiv preprint arXiv:2207.11652}, 2022.

\bibitem{qian2021counterfactual}
Chen Qian, Fuli Feng, Lijie Wen, Chunping Ma, and Pengjun Xie.
\newblock Counterfactual inference for text classification debiasing.
\newblock In {\em Proceedings of the 59th Annual Meeting of the Association for Computational Linguistics and the 11th International Joint Conference on Natural Language Processing (Volume 1: Long Papers)}, pages 5434--5445, 2021.

\bibitem{yang2024towards}
Dingkang Yang, Mingcheng Li, Dongling Xiao, Yang Liu, Kun Yang, Zhaoyu Chen, Yuzheng Wang, Peng Zhai, Ke~Li, and Lihua Zhang.
\newblock Towards multimodal sentiment analysis debiasing via bias purification.
\newblock {\em arXiv preprint arXiv:2403.05023}, 2024.

\bibitem{pearl2009causality}
Judea Pearl.
\newblock {\em Causality}.
\newblock Cambridge University Press, 2009.

\bibitem{niu2021counterfactual}
Yulei Niu, Kaihua Tang, Hanwang Zhang, Zhiwu Lu, Xian-Sheng Hua, and Ji-Rong Wen.
\newblock Counterfactual vqa: A cause-effect look at language bias.
\newblock In {\em Proceedings of the IEEE/CVF Conference on Computer Vision and Pattern Recognition (CVPR)}, pages 12700--12710, 2021.

\bibitem{yang2023context}
Dingkang Yang, Zhaoyu Chen, Yuzheng Wang, Shunli Wang, Mingcheng Li, Siao Liu, Xiao Zhao, Shuai Huang, Zhiyan Dong, Peng Zhai, and Lihua Zhang.
\newblock Context de-confounded emotion recognition.
\newblock In {\em Proceedings of the IEEE/CVF Conference on Computer Vision and Pattern Recognition (CVPR)}, pages 19005--19015, June 2023.

\bibitem{pearl2000models}
Judea Pearl et~al.
\newblock Models, reasoning and inference.
\newblock {\em Cambridge, UK: CambridgeUniversityPress}, 19:2, 2000.

\bibitem{xu2015show}
Kelvin Xu, Jimmy Ba, Ryan Kiros, Kyunghyun Cho, Aaron Courville, Ruslan Salakhudinov, Rich Zemel, and Yoshua Bengio.
\newblock Show, attend and tell: Neural image caption generation with visual attention.
\newblock In {\em International Conference on Machine Learning (ICML)}, pages 2048--2057. PMLR, 2015.

\bibitem{le2015tiny}
Ya~Le and Xuan Yang.
\newblock Tiny imagenet visual recognition challenge.
\newblock {\em CS 231N}, 7(7):3, 2015.

\bibitem{simonyan2014very}
Karen Simonyan and Andrew Zisserman.
\newblock Very deep convolutional networks for large-scale image recognition.
\newblock {\em arXiv preprint arXiv:1409.1556}, 2014.

\bibitem{zagoruyko2016wide}
Sergey Zagoruyko and Nikos Komodakis.
\newblock Wide residual networks.
\newblock {\em arXiv preprint arXiv:1605.07146}, 2016.

\bibitem{zhang2017mixup}
Hongyi Zhang, Moustapha Cisse, Yann~N Dauphin, and David Lopez-Paz.
\newblock mixup: Beyond empirical risk minimization.
\newblock {\em arXiv preprint arXiv:1710.09412}, 2017.

\bibitem{zhao2022decoupled}
Borui Zhao, Quan Cui, Renjie Song, Yiyu Qiu, and Jiajun Liang.
\newblock Decoupled knowledge distillation.
\newblock In {\em Proceedings of the IEEE/CVF Conference on Computer Vision and Pattern Recognition (CVPR)}, pages 11953--11962, 2022.

\bibitem{ioffe2015batch}
Sergey Ioffe and Christian Szegedy.
\newblock Batch normalization: Accelerating deep network training by reducing internal covariate shift.
\newblock In {\em International Conference on Machine Learning (ICML)}, pages 448--456. pmlr, 2015.

\bibitem{van2008visualizing}
Laurens Van~der Maaten and Geoffrey Hinton.
\newblock Visualizing data using t-sne.
\newblock {\em Journal of Machine Learning Research}, 9(11), 2008.

\bibitem{paszke2019pytorch}
Adam Paszke, Sam Gross, Francisco Massa, Adam Lerer, James Bradbury, Gregory Chanan, Trevor Killeen, Zeming Lin, Natalia Gimelshein, Luca Antiga, et~al.
\newblock Pytorch: An imperative style, high-performance deep learning library.
\newblock {\em Advances in Neural Information Processing Systems (NeurIPS)}, 32, 2019.

\bibitem{binici2022preventing}
Kuluhan Binici, Nam~Trung Pham, Tulika Mitra, and Karianto Leman.
\newblock Preventing catastrophic forgetting and distribution mismatch in knowledge distillation via synthetic data.
\newblock In {\em Proceedings of the IEEE/CVF Winter Conference on Applications of Computer Vision (WACV)}, pages 663--671, 2022.

\bibitem{binici2022robust}
Kuluhan Binici, Shivam Aggarwal, Nam~Trung Pham, Karianto Leman, and Tulika Mitra.
\newblock Robust and resource-efficient data-free knowledge distillation by generative pseudo replay.
\newblock In {\em Proceedings of the AAAI Conference on Artificial Intelligence (AAAI)}, volume~36, pages 6089--6096, 2022.

\bibitem{patel2023learning}
Gaurav Patel, Konda~Reddy Mopuri, and Qiang Qiu.
\newblock Learning to retain while acquiring: Combating distribution-shift in adversarial data-free knowledge distillation.
\newblock In {\em Proceedings of the IEEE/CVF Conference on Computer Vision and Pattern Recognition (CVPR)}, pages 7786--7794, 2023.

\end{thebibliography}
}


\newpage

\maketitlesupplementary

\begin{figure*}[h]
	\centering
    %\vspace{-0.1cm}
	\includegraphics[scale=0.45]{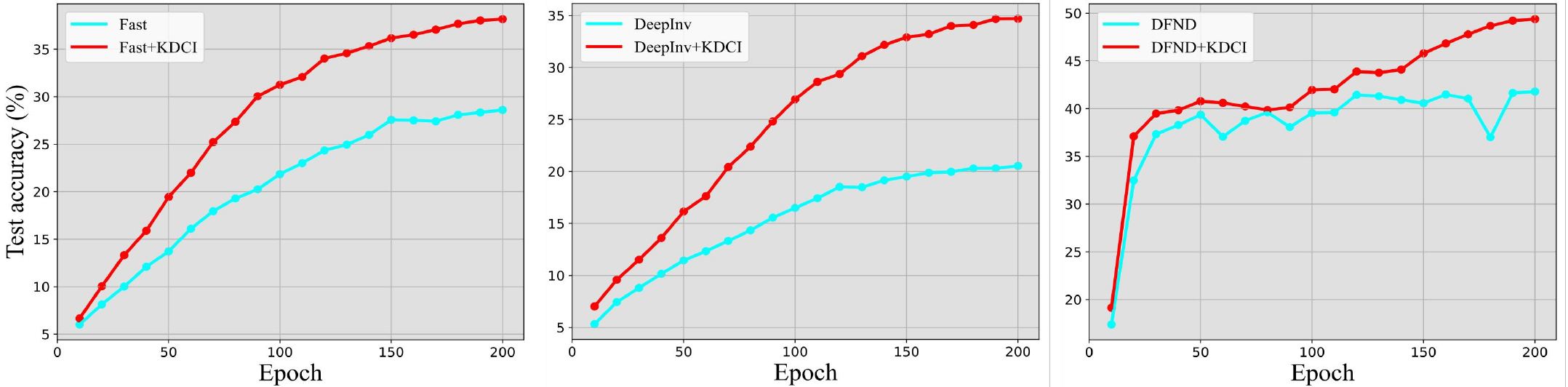}
	%\vspace{-0.5cm}
	\caption{The test accuracy on Tiny-ImageNet dataset across different local training epochs $E=\left \{ 10,20,\dots,200\right \} $. Our KDCI framework improves the performance of baselines consistently.}
	\label{fig1_sup}
	%\vspace{-0.1cm}
\end{figure*}

In this supplementary material, we provide more details
of our method, organized as follows:

\begin{itemize}
    \item In Section~\ref{secA}, we provide the detailed training settings and illustrate how KDCI combines with existing DFKD methods, and show the algorithm process, corresponding to Section \textcolor{blue}{4.3} of the main body.

    \item In Section~\ref{secB}, we qualitatively assess students' learning progress about vanilla DFKD methods and their KDCI-based version to verify the positive effect of KDCI on the existing DFKD method.
    
    \item In Section~\ref{secC}, we analyze the possible reasons for the difference in performance improvement, corresponding to Section \textcolor{blue}{4} of the main body.

    \item In Section~\ref{secD}, we provide more observable visualization results as more sufficient evidence, corresponding to Section \textcolor{blue}{4.7} of the main body.

    \item In Section~\ref{secE}, we discuss the significant differences between our KDCI and other methods focusing on data distribution.

    \item In Section~\ref{secF}, we discuss the broader impact and potential limitations.

    \item In Section~\ref{secG}, we provide the detailed experimental settings for the used baseline methods, corresponding to Section \textcolor{blue}{4.2} of the main body.

\end{itemize}

\begin{algorithm}[th]

  \begin{algorithmic}[1]
  \caption{Training process of generation-based methods combined with our KDCI}
  \label{alg1}
    \small{
    \Require
        A pre-trained teacher model $\bm{T}$, a generator $g$, a student model $\bm{S}$, distillation epochs $T$, batch size $N_m$, the iterations of generator $g$ in each epoch $T\!g$, the iterations of student $f_{s}$ in each epoch $T\!s$, the confounder size $N$.
    
    \For {epoch$\;= [1,\dots,T]$}  
        \State $//$ \textit{\textbf{Generation stage}}
        \For {generator iterations $\;= [1,\dots,T\!g]$}  
        \State Randomly sample noises and labels $({\textrm{z}},{y})$ 
        \State Synthesize a mini-batch training data $\bm{X}=g(\textrm{z},{{y}})$
        \State Update generator $g$ with the generator loss
        
        \EndFor

        \State Synthesize training data $\bm{X}=g(\textrm{z},{{y}})$. Obtain the predic-
        \Statex \;\;\;\; tion feature $M = \left \{  m_j \in \mathbb{R}^d  \right \}_{j=1}^{N_m}$
        \State Prototype clustering for $M$. Calculate the number of the 
        \Statex \;\;\;\; prediction features in $i$-th cluster 
        \Statex \;\;\;\; $N_i$, the feature cluster $ {\textstyle \sum_{k=1}^{N_i}} m_k^i$ and the subcenter 
        \Statex \;\;\;\; $\bm{z}_i =  \frac{1}{N_i} {\textstyle \sum_{k=1}^{N_i}} m_k^i$.
        \State Construct a confounder dictionary $\bm{Z}=\left [ \bm{z}_1,\bm{z}_2, \dots, \bm{z}_{N} \right ]$ 
        \Statex \;\;\;\; and calculate the prototype proportion $P_s(\bm{z}_i)=N_i/N_m$

        \State $//$ \textit{\textbf{Distillation stage}}
        \For {student iterations $\;= [1,\dots,T\!s]$}  

        \State Synthesize training data $\bm{X}=g(\textrm{z},{{y}})$. Get models's
        \Statex \;\;\;\;\;\;\;\;\;\; predictions $\bm{T}(\bm{X})$ and $\bm{S}(\bm{X})$
        \State Calculate the prior information:
        \Statex \;\;\;\;\;\;\;\;\;\; $F(\bm{z})=\sum_{i=1}^{N}\lambda_i \bm{z}_i P_s(\bm{z}_i)$
        \State Compensate the student's predictions:
        \Statex \;\;\;\;\;\;\;\;\;\; $\bm{S'}(\bm{X})= \phi(\bm{S}(\bm{X}), F(\bm{z}) ) $
        \State Update the student $\bm{S}$ with $K\!D\langle \bm{T}(\bm{X}), \bm{S'}(\bm{X}) \rangle$
        \EndFor
    \EndFor
    \Ensure The student model $\bm{S}$.
    }
    \end{algorithmic}

\end{algorithm}

\renewcommand\thesection{\Alph{section}}
\setcounter{section}{0}

%\section{Section A: Algorithms of existing DFKD methods combined with our KDCI}
\section{Additional Training Details \& Algorithm Process of Combining KDCI with Existing DFKD Methods}
\label{secA}

\subsection{Training Details}
We provide the detailed experimental settings for our KDCI framework.
Our KDCI and reproducible methods are implemented through PyTorch \cite{paszke2019pytorch}.
All models are trained on RTX 3090 GPUs.
\textbf{For CIFAR-10 and CIFAR-100}, all training settings (\textit{e.g.}, loss function, optimizer, batch size, learning rate, etc) of the reported methods are consistent with the released codebase. The results are shown in Table~1 of the main body.
\textbf{For Tiny-ImageNet}, initially, we try to find a unified teacher model for the Tiny-ImageNet dataset in open-sourced projects. However, one problem is that the teacher model pre-trained on Tiny-ImageNet seems confidential, so finding an open-source unified model is difficult.
In this case, we train the unified renset-34 teacher model for 200 epochs on the original training data.
During the teacher's training, we use the SGD optimizer with the momentum as 0.9, weight decay as 5$e-$4, the batch size as 128, and cosine annealing learning rate with an initial value of 0.1.
The teacher model can converge without additional tuning.
Based on this pre-trained teacher, we train all students for 200 epochs.
For the student, we use the SGD optimizer with the momentum as 0.9, the weight decay as 1$e-$4, the batch size as 256, the cosine annealing learning rate with an initial value of 0.2 for Fast \cite{fang2022up}, and 0.1 for DeepInv \cite{bhardwaj2019dream} \& DFND \cite{chen2021learning}.
The results are shown in Table~2 of the main body.
\textbf{For ImageNet}, We choose the same pre-trained resnet-50 model with \cite{zhao2022decoupled} and unify the teacher model of different baseline methods.
For Fast, we test directly on the open-source project.
For DeepInv, we reproduce the corresponding results with the specified backbone pair.
For DFND, we select 600k samples from the unlabeled FlickerlM dataset. 
The teacher's backbone is different from the original paper. 
The different backbones may cause the results we reproduce to differ from the original paper.
The results are shown in Table~1 of the supplementary material.
For the implementation of our KDCI, the hidden dimension $d_n$ is set to 256.
And $d_h$ equals the hidden dimension $d$ and the number of classes.
By default, $\phi(\cdot)$ uses feature addition.
For various baseline methods, the settings are shown in Section~\ref{secG} of the supplementary material.

\subsection{Algorithm Process}
In the existing DFKD task, the generation-based and sampling-based method processes are different.
Therefore, the way KDCI combines these methods and the hyperparameter settings are also slightly different.
For the generation-based process, the generator and student models are updated alternately, which means the student's training data is updated in each epoch.
We use a mini-batch of synthetic training data to construct the confounder dictionary, and the dictionary will be updated as the generator is updated.
For the sampling-based process, existing methods select unlabeled data according to the preferences of the teacher model.
Then, the student relies on these unlabeled data for data-based knowledge distillation training.
We use all sampled data to construct the confounder dictionary.
During subsequent student training, the dictionary is fixed.
For a clearer understanding, we describe the above process as Algorithm~\ref{alg1} and \ref{alg2}, respectively.

%\section{Section B: Vanilla DFKD methods vs. their KDCI-based version}
\section{Vanilla DFKD Methods vs. Their KDCI-based Versions}
\label{secB}

In the main body, we have compared the quantitative results of vanilla DFKD methods and their KDCI-based versions.
To observe the positive effect of KDCI on the existing DFKD methods more clearly, we visualize the student's test accuracy on the Tiny-ImageNet dataset.
The results are shown in Figure~\ref{fig1_sup}.
KDCI can consistently help students from the beginning of training to the end, which verifies its effectiveness.

\begin{algorithm}[t]

  \begin{algorithmic}[1]
  \caption{Training process of sampling-based methods combined with our KDCI}
  \label{alg2}
  \small{
    \Require
        A pre-trained teacher model $\bm{T}$, a student model $\bm{S}$, unlabeled training dataset $D=\left \{ x_j \right \}_{j=1}^{n}$, distillation epochs $T$, batch size $m$, number of batches $M$, the number of sampled data $N_m$, the confounder size $N$.
    
    %\State Initialize parameter $\theta$
    \State $//$ \textit{\textbf{Sampling stage}}
    \State Sample the training data $\left \{ x_j \right \}_{j=1}^{N_m}$ from $D$. Obtain the prediction feature set $M = \left \{  m_j \in \mathbb{R}^d  \right \}_{j=1}^{N_m}$
    \State Prototype clustering for $M$. Calculate the number of the prediction features in $i$-th cluster $N_i$, the feature cluster $ {\textstyle \sum_{k=1}^{N_i}} m_k^i$ and the subcenter $\bm{z}_i =  \frac{1}{N_i} {\textstyle \sum_{k=1}^{N_i}} m_k^i$.
    \State Construct a confounder dictionary $\bm{Z}=\left [ \bm{z}_1,\bm{z}_2, \dots, \bm{z}_{N} \right ]$ and calculate the prototype proportion $P_s(\bm{z}_i)=N_i/N_m$

    \State $//$ \textit{\textbf{Distillation stage}}
    \For {epoch$\;= [1,\dots,T]$}  
        \For {mini-batch$\;= [1,\dots,M]$}
        \State Sample a mini-batch training data:
        \Statex \;\;\;\;\;\;\;\;\;\; $\bm{X}= \left \{ x_i \right \}_{i=1}^{m}$ from $\left \{ x_j \right \}_{j=1}^{N_m}$
        \State Get teacher and student predictions $\bm{T}(\bm{X})$ and $\bm{S}(\bm{X})$
        \State Calculate the prior information:
        \Statex \;\;\;\;\;\;\;\;\;\; $F(\bm{z})=\sum_{i=1}^{N}\lambda_i \bm{z}_i P_s(\bm{z}_i)$
        \State Compensate the student's predictions:
        \Statex \;\;\;\;\;\;\;\;\;\; $\bm{S'}(\bm{X})= \phi(\bm{S}(\bm{X}), F(\bm{z}) ) $
        \State Update the student $\bm{S}$ with $K\!D\langle \bm{T}(\bm{X}), \bm{S'}(\bm{X}) \rangle$
        \EndFor
    \EndFor
    \Ensure The student model $\bm{S}$.
    }
    \end{algorithmic}
\end{algorithm}

\section{Analyses of Difference in Performance Improvements}
\label{secC}

Judging from the experimental results, KDCI has different gains for different DFKD methods on different datasets. We think such observations arise from various factors.
\begin{itemize}
    \item By default, we choose the teacher model itself to extract the confounding dictionary. The prediction feature set provided by teachers of different backbones has different expressiveness, which affects the compensation degree of backdoor adjustment for bias during the causal intervention. The tests in Lines 513-531 and Table. 5 of the main body also verify this conclusion.
    \item The degree of distribution shift of synthetic data on distinct datasets is different. More complex datasets may degrade the generation quality for generation-based methods, resulting in more significant distribution shifts. KDCI tends to be more effective for more sophisticated datasets.
    \item Different baseline methods with different training losses are influential. Observations such as Section \textcolor{blue}{4.4} of the main body suggest that methods that already incorporate prior likelihood knowledge of the data may weaken the KDCI gain.
    \item In addition, there may be many underlying factors. Nevertheless, KDCI, as a model-agnostic general framework, has promising and competitive improvements and gains for various models as a whole. 
    We believe that a deeper exploration of the relevant mechanisms is a promising perspective. For this topic, we leave it to future work.
\end{itemize}

\begin{figure}[h]
	\centering
    %\vspace{-0.1cm}
	\includegraphics[scale=0.563]{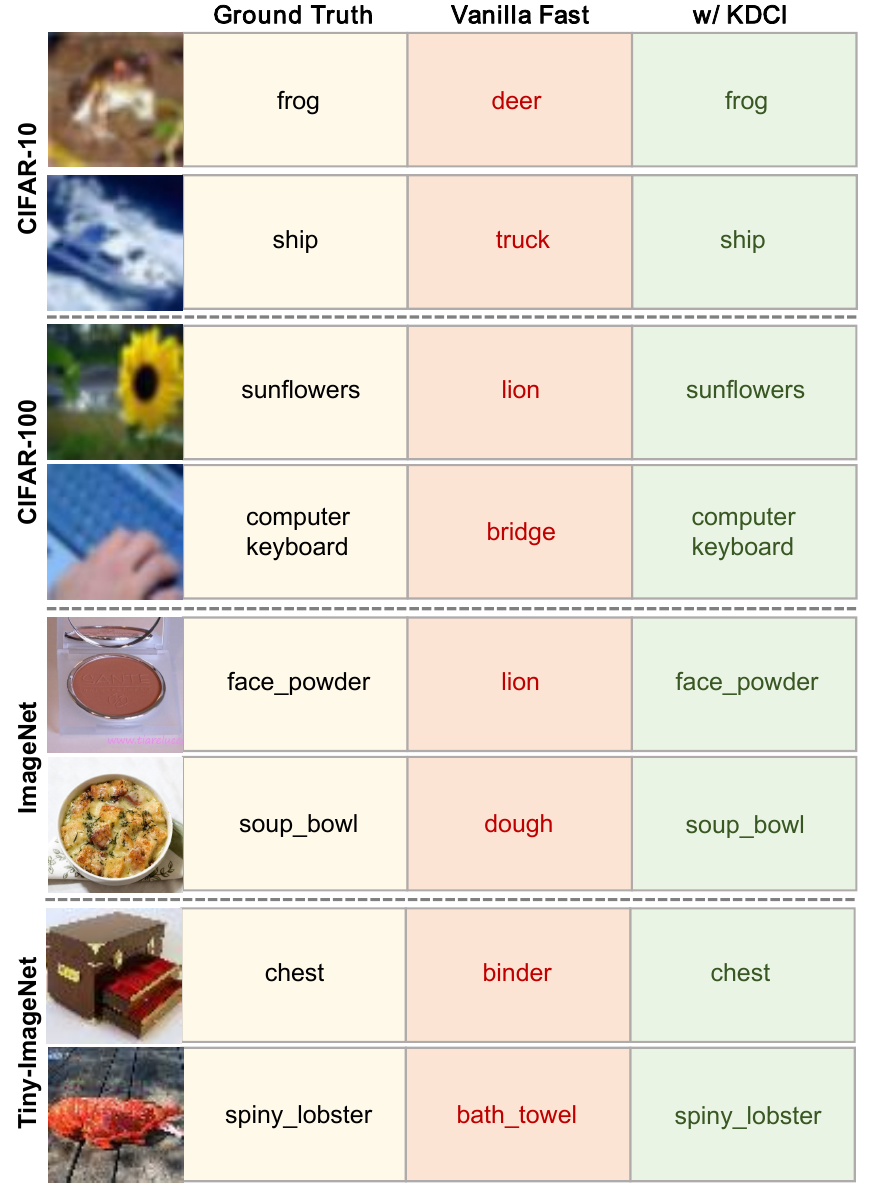}
	%\vspace{-0.5cm}
	\caption{Qualitative results of the vanilla and KDCI-based version on CIFAR-10, CIFAR-100, ImageNet, and Tiny-ImageNet.}
	\label{fig2_sup}
	%\vspace{-0.1cm}
\end{figure}

\section{More Visual Evidence}
\label{secD}

To further verify the effectiveness, we provide more case studies of causal intervention.
As shown in Figure~\ref{fig2_sup}, we visualize some test instances corrected by our KDCI compared to the vanilla version (Fast) on four kinds of datasets (\ie, CIFAR-10, CIFAR-100, ImageNet, and Tiny-ImageNet).
The vanilla version sometimes confuses some test instances due to shape or color.
Our KDCI can repair these prediction shifts to enhance student performance.

\section{Discussion with Other Works that Address Distribution Shifts}
\label{secE}

Several DFKD works already address distribution shifts in adversarial contexts \cite{binici2022preventing, binici2022robust, do2022momentum, patel2023learning}.
The works reveal distribution shift issues in the DFKD task from different aspects, but our method is significantly different from these works.
Specifically, the differences between our KDCI and others are as follows:
\begin{itemize}
\item  \textbf{Applicability.} These existing works tacitly use the same motivation, \textit{i.e.}, as the generator gets updated, the distribution of synthetic data will change, causing the student to forget the knowledge it acquired at previous steps. However, such motivation does not apply to sampling-based methods. After selecting the training samples, they will not change during the entire student training process. Our motivation comes from the observed distribution shifts between the substitution data and can cover the two methods mentioned.
\item \textbf{Economy.} Existing methods often rely on substantial additional computational and storage costs, \textit{e.g.}, the need to store and maintain an additional dynamic collection of generated samples \cite{binici2022preventing}, the need for additional generator architectures to memorize knowledge of past generated data (an additional Variational Autoencoder (VAE) \cite{binici2022robust} or Exponential Moving Average generator \cite{do2022momentum}), and additional memory bank or additional loss calculation and gradient update \cite{patel2023learning}. In contrast, our method only needs to compute and store a small number of matrix computation results. Compared with the update of the models, the computational cost of the clustering process is basically negligible.
\item \textbf{Plug-and-play.} Existing works are to propose new methods. Undoubtedly, these methods can provide a potential reference for other DFKD methods, but whether they can be easily combined with existing DFKD methods and improve overall performance is still unknown. Our proposed technique is model-agnostic, as a plug-and-play paradigm that integrates well with existing works. A large number of experiments have proved this conclusion.
\end{itemize}

\section{Further Discussion}
\label{secF}

\subsection{Broader Impact}
The positive impact of this work: 
the proposed KDCI module can suppress the distribution shifts between the substitution and original data in the DFKD task, preventing the potential discrimination of the student's learning.
While the pre-trained model for extracting prior knowledge uses the teacher itself, our method does not require additional dependencies and auxiliary information.
The negative impacts of this work: 
students may be forced to identify minority groups for malicious purposes with customized biased teacher models.
Therefore, we have to make sure that the DFKD technique is used for the right purpose.

\subsection{Limitations}
Since there are countless methods with insights for the DFKD task, other ways of classifying forms may also be reasonable.
In this paper, we simply divide the source of the substitution data into generation-based and sampling-based methods.
Similarly, it is impossible to cover all DFKD methods, so only open-source and representative methods are selected as the baseline.
Nevertheless, the existing performance improvement is enough to prove the positive impact of KDCI on students.

In addition, since what we propose is a framework rather than a specific method, the test on the effectiveness of KDCI relies on the experimental setting of the existing DFKD methods.
Currently, the mainstream open-source DFKD methods rarely use real-life medical or facial datasets for testing, so we only follow the mainstream experimental settings.
Following the consensus of peers is necessary to increase the impact of our work. In this work, we select datasets that are widely used and accepted by the vast majority of DFKD methods. Following previous data paradigms is beneficial for acceptance by the relevant research community and enhances the persuasiveness of our method.

\section{Experimental Setup of the Baseline DFKD Methods}

\label{secG}

\textbf{DAFL.}
DAFL~\cite{chen2019data} is a data-free generation method.
We keep the generator loss from the original as: $\mathcal{L}_{GEN}=\mathcal{L}_{oh}+ \alpha\mathcal{L}_{a}+ \beta\mathcal{L}_{ie} $.
The knowledge distillation loss is:
$\mathcal{L}_{KD}=D_{KL}(\mathcal{N}_S(x),\mathcal{N}_T(x)) $.
Following the original settings, we set $\alpha=1e-3$, $\beta=20$. We use SGD with the weight decay of $5e-4$, the momentum of $0.9$, and the initial learning rate set as $0.1$.

\noindent \textbf{Fast.}
Fast \cite{fang2022up} is a fast data-free generation method via feature sharing.
We keep the generator loss from the original as: $ \mathcal{L}_{GEN} = \alpha\mathcal{L}_{cls} + \beta\mathcal{L}_{adv} + \gamma\mathcal{L}_{feat}$.
The knowledge distillation loss is:
$\mathcal{L}_{KD}=D_{KL}(\mathcal{N}_S(x),\mathcal{N}_T(x)) $.
We set $\alpha=0.4$, $\beta=1.1 $, and $\gamma=10$, which are the same as the original settings.
We use the Adam Optimizer with a learning rate of $1e-3$ to update the generator and the SGD optimizer with a momentum of $0.9$ and a learning rate of $0.1$ for student training.

\noindent \textbf{CMI.}
CMI~\cite{fang2021contrastive} is a model inversion method with contrastive learning.
We keep the generator loss from the original as: $ \mathcal{L}_{GEN} = \alpha\mathcal{L}_{bn}+ \beta\mathcal{L}_{cls} + \gamma\mathcal{L}_{adv} + \delta\mathcal{L}_{cr}$.
The knowledge distillation loss is:
$\mathcal{L}_{KD}=D_{KL}(\mathcal{N}_S(x),\mathcal{N}_T(x)) $.
We set $\alpha=1$, $\beta=0.5 $, $\gamma=0.5 $, and $\delta=0.8$.
We use the Adam Optimizer with a learning rate of $1e-3$ to update the generator and the SGD optimizer with a momentum of $0.9$ and a learning rate of $0.1$ for student training.

\noindent \textbf{DeepInv.}
DeepInv~\cite{yin2020dreaming} is a model inversion method that combines prior knowledge and adversarial training.
We keep the inversion loss from the original as: $\mathcal{L}_{GEN}=\alpha_{tv}\mathcal{R}_{tv}+ \alpha_{l2} \mathcal{R}_{l2} + \alpha_{f} \mathcal{R}_{feature} + \alpha_{c}\mathcal{R}_{compete}$.
The knowledge distillation loss is:
$\mathcal{L}_{KD}=D_{KL}(\mathcal{N}_S(x),\mathcal{N}_T(x)) $.
We set $\alpha_{tv}=2.5e-5$ , $\alpha_{l2}=3e-8$, $\alpha_{f}=0.1$ and $\alpha_{c}=10$, which are the same as the original setting.
Besides, we set the number of iterations as $1000$ and use Adam for optimization with a learning rate of $0.05$.

\noindent \textbf{DFND.}
DFND~\cite{chen2021learning} is a sampling-based method using open-world unlabeled data as the substitution data.
Following the original, we select 600k data with the highest teacher confidence from the ImageNet dataset \cite{deng2009imagenet} as the sampled data and resize them to the resolution of the corresponding dataset.
We use the same noisy distillation loss $ \mathcal{L}_{KD}=\mathcal{H}_{CE}(Q(\mathcal{N}_S(x)),\hat{y} ) +\lambda D_{KL}(\mathcal{N}_S(x),\mathcal{N}_T(x))$, and $\lambda$ is set as 4.
The student network is optimized using SGD and the initial learning rate is set as $0.1$
Weight decay and momentum are set as $5e-4$ and $0.9$, respectively.

\noindent \textbf{Mosaick.}
Mosaick~\cite{fang2021mosaicking} is a sampling-based method using out-of-domain (OOD) unlabeled data as the substitution data.
We select 600k data with the lowest teacher confidence from the ImageNet dataset \cite{deng2009imagenet} as the OOD data.
Following the original settings, 
we use Adam for optimization, with hyper-parameters {$lr=1e-3$, $\beta_1=0.5 $, and $\beta_2=0.999 $} for the generator and discriminator.
The distillation loss is $\mathcal{L}_{KD}= \lambda D_{KL}-\lambda\mathcal{R}(G,D,T) $
The student network is optimized using SGD, and the initial learning rate is set as $0.1$.
Weight decay and momentum are set as $1e-4$ and $0.9$, respectively.

\end{document}